\documentclass{article}

\usepackage[preprint]{neurips_2025}

\usepackage[utf8]{inputenc} 
\usepackage[T1]{fontenc}    
\usepackage{hyperref}       
\usepackage{url}            
\usepackage{booktabs}       
\usepackage{amsfonts}       
\usepackage{nicefrac}       
\usepackage{microtype}      
\usepackage{xcolor}         
\usepackage{authblk}
\usepackage{graphicx}
\usepackage{amsmath, amssymb}
\usepackage{xcolor}         
\usepackage{tikz}
\usepackage[edges]{forest}
\usepackage{makecell}

\title{Scientific Hypothesis Generation and Validation: Methods, Datasets, and Future Directions}

\author[1]{Adithya Kulkarni}
\author[1]{Fatimah Alotaibi}
\author[1]{Xinyue Zeng}
\author[1]{Longfeng Wu}
\author[1]{Tong Zeng}
\author[1]{Barry Menglong Yao}
\author[1]{Minqian Liu}
\author[1]{Shuaicheng Zhang}
\author[1]{Dawei Zhou}
\author[2]{Lifu Huang\thanks{Correspondence to: \texttt{aditkulk@vt.edu}, \texttt{lfuhuang@ucdavis.edu}}}

\affil[1]{Virginia Tech, Blacksburg, VA, USA}
\affil[2]{University of California, Davis, CA, USA}

\begin{document}

\maketitle

\section{Introduction}

Scientific discovery has long been the subject of computational modeling, particularly through systems that frame discovery as a structured problem-solving process grounded in cognitive science and artificial intelligence~\cite{bradshaw1983studying, simon1992scientific, langley1988computational, langley1998computer, langley2000computational, dvzeroski2007computational, langley2013central, langley2024integrated, Bengio+chapter2007, Hinton06, BODEN1998347}. These early approaches emphasized the iterative formulation and refinement of hypotheses using heuristic rules, domain knowledge, and symbolic representations. While such systems successfully simulated elements of scientific reasoning, ranging from rediscovering physical laws to inferring causal structures, their scalability and adaptability to unstructured data remained limited.
In parallel, the advent of Large Language Models (LLMs)~\cite{llama3modelcard, openai2024gpt4ocard} has marked a transformative shift in scientific knowledge creation and validation. These models, trained on expansive corpora encompassing text, numerical data, and multimodal inputs, possess the remarkable capability to synthesize diverse datasets, identify latent patterns, and accelerate the hypothesis generation process at an unprecedented scale~\cite{beltagy2019scibert, qi2024large, wang2023scimon, WANG2025100007, doi:10.1126/sciadv.adr9038, 10765518, goodfellow2016deep}. Moreover, LLMs have demonstrated proficiency in extracting meaningful relationships from unstructured text, facilitating hypothesis discovery in fields such as biomedical research and materials science~\cite{sybrandt2018large, ghafarollahi2024protagents, doi:10.1126/sciadv.adu4391, bran2023chemcrow, kadic20193d, bertoldi2017flexible, jia2020engineering, jiao2021artificial, bauer2017nanolattices, papadimitriou2024ai}. These capabilities make LLMs instrumental in confronting the complexity and scale of contemporary scientific inquiry, enabling a paradigm where data-driven insights complement and extend human reasoning in the discovery process~\cite{yang2023large, shojaee2024llm, romera2024mathematical, trinh2024solving, lesica2021harnessing}.
\begin{figure*}
    \centering
    \includegraphics[width=\linewidth]{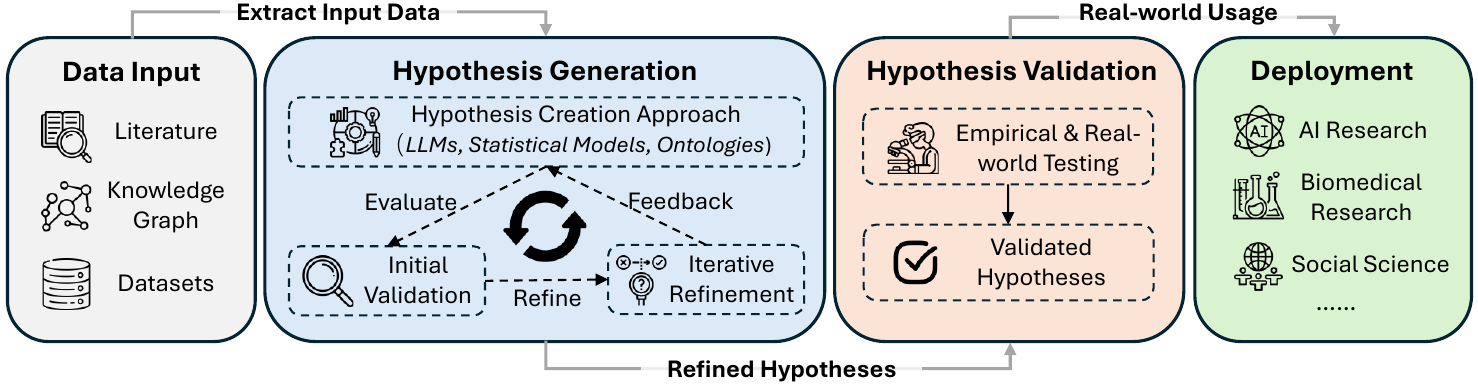}
    \caption{Overview of the scientific hypothesis generation and validation pipeline integrating LLMs, statistical models, and ontologies. The figure illustrates the stages from data input and hypothesis creation to iterative validation and real-world deployment, highlighting the feedback loops that refine hypotheses over time.}
    \label{fig:figure_1}
\end{figure*}

The growing capabilities of LLMs underscore their potential to revolutionize hypothesis generation and validation. Frameworks such as The AI Scientist~\cite{lu2024ai} and SciAgents~\cite{ghafarollahi2024sciagents} epitomize the advancements in agentic AI systems that autonomously undertake significant elements of the scientific process, including experimental validation and the drafting of manuscripts. These systems harness advanced methodologies such as retrieval-augmented generation (RAG)~\cite{chen2024chemist}, knowledge graph integration~\cite{zhou2024hypothesisgenerationlargelanguage, sybrandt2017moliere}, and causal inference~\cite{jha2019hypothesis}, enabling the generation of hypotheses that are not only testable but also interdisciplinary. By systematically mapping connections across seemingly unrelated domains, LLMs uncover insights that human researchers might overlook due to cognitive constraints or disciplinary silos~\cite{fawzi2022discovering, touvron2023llama, socraticqa}. This integration of machine learning models and LLMs into the hypothesis generation process redefines the boundaries of scientific exploration, opening avenues for cross-domain innovations that were previously inconceivable. Figure~\ref{fig:figure_1} provides an overview of the scientific hypothesis generation and validation pipeline.

\begin{table}[ht]
\centering
\caption{Examples of LLM-Driven Tools for Hypothesis Creation and Validation}
\begin{tabular}{p{4cm}|p{4cm}|p{4.1cm}}
\hline
\textbf{Tool} & \textbf{Application} & \textbf{Contribution to Scientific Discovery} \\ \hline
\textit{AlphaFold}~\cite{jumper2021highly} & Protein structure prediction & Resolved a decades-long challenge in biology, accelerating drug discovery \\ \hline
\textit{Crispr-GPT}~\cite{huang2024crispr} & Gene-editing experiment design & Automated hypothesis creation for CRISPR-based research \\ \hline
\textit{SciAgents}~\cite{ghafarollahi2024sciagents} & Dynamic knowledge graph generation & Maps relationships between interdisciplinary concepts, revealing unexplored connections \\ \hline
\textit{Discovering Faster Matrix Multiplication Algorithms}~\cite{fawzi2022discovering} & Algorithmic optimization & Demonstrates how reinforcement learning can refine complex mathematical hypotheses \\ \hline
\textit{Materials Project}~\cite{jain2020materials} & Materials property prediction & Enables hypothesis generation about novel material structures and properties \\ \hline
\textit{MOLIERE}~\cite{sybrandt2017moliere, sybrandt2018large} & Biomedical hypothesis validation & Retrospectively tests hypotheses against historical biomedical data and identifies novel biomedical insights \\ \hline
\end{tabular}
\label{tab:examples_tools}
\end{table}

The transformative potential of LLMs is further exemplified by their applications in addressing long-standing scientific challenges. For instance, AlphaFold \cite{jumper2021highly} has revolutionized protein structure prediction, resolving key bottlenecks in drug discovery and expediting therapeutic innovation. Similarly, Crispr-GPT \cite{huang2024crispr} streamlines the design of gene-editing experiments, reducing the cognitive and procedural burdens on researchers while accelerating the pace of scientific advancement. In addition, MOLIERE \cite{sybrandt2017moliere, sybrandt2018large}demonstrates how text mining and biomedical knowledge graphs can aid hypothesis validation by retrospectively testing hypotheses against historical data. These tools underscore the dual role of LLMs in augmenting human capabilities and enabling breakthroughs that transcend traditional boundaries of scientific inquiry. Table~\ref{tab:examples_tools} provides a comprehensive overview of notable tools and their contributions to hypothesis generation and validation, illustrating the breadth of impact that these systems have achieved.

Despite their remarkable capabilities, LLMs exhibit fundamental differences from human researchers, particularly in reasoning and knowledge synthesis. Human cognition is characterized by inherent intuition, creativity, and contextual understanding~\cite{langley2013central}, enabling the pursuit of unconventional pathways that often lead to groundbreaking discoveries \cite{fawzi2022discovering}. Conversely, LLMs operate within the probabilistic framework of pattern recognition, constrained by the biases and limitations of their training datasets \cite{tari2010discovering}. While this allows for efficient data processing and knowledge reinforcement, it often results in the perpetuation of established ideas rather than the generation of genuinely novel concepts. This divergence underscores the necessity of harmonizing human intuition with machine-driven capabilities to achieve meaningful progress in scientific discovery \cite{fok2024search, zhou2024hypothesisgenerationlargelanguage, wang2023scimon}. The success of agentic AI frameworks like SciAgents \cite{ghafarollahi2024sciagents} further demonstrates how hybrid human-AI collaboration can refine hypotheses, ensuring that machine-generated insights remain aligned with human reasoning. Such integration emphasizes the complementary roles of humans and machines, with each compensating for the limitations of the other.

Agentic AI systems~\cite{paul2024agentpeertalkempoweringstudentsagenticaidriven, white2024buildinglivingsoftwaresystems, 10.1145/3593013.3594033, sulc2024agenticaiparticleaccelerators, qiu2024interactive}, powered by Large Language Models (LLMs), are reshaping the landscape of scientific discovery by automating routine but essential tasks such as data analysis, hypothesis formulation, and literature synthesis. These systems allow researchers to redirect their cognitive resources toward more creative and complex endeavors, thereby augmenting human ingenuity rather than replacing it. A recent comprehensive survey~\cite{gridach2025agentic} provides a detailed taxonomy of agentic systems, distinguishing between autonomous and collaborative frameworks and categorizing their deployment across domains like chemistry, biology, and materials science. The study highlights their role across the full research lifecycle—from ideation and literature review to experimentation and scientific writing—and emphasizes human-AI collaboration and system calibration as pivotal directions for future development. Notable tools such as SciAgents~\cite{ghafarollahi2024sciagents}, AI Co-Scientist~\cite{gottweis2025towards}, and reinforcement learning-driven frameworks for materials discovery~\cite{gruver2024fine} exemplify how domain-specific repositories and real-time feedback loops can enhance the contextual relevance and applicability of generated hypotheses. Likewise, systems like Discovering Faster Matrix Multiplication Algorithms~\cite{fawzi2022discovering} and causal inference models in biomedical research~\cite{jha2019hypothesis} further demonstrate the versatility of LLM-integrated scientific agents. However, despite these advancements, challenges remain, particularly concerning over-reliance on pre-existing data, limited novelty in generated hypotheses, and ethical considerations in high-stakes domains like healthcare~\cite{tang2024prioritizing, shavit2023practices}. These limitations underscore the need for responsible, human-in-the-loop designs that balance automation with domain expertise and ethical oversight.

This survey distinguishes itself by offering a holistic, interdisciplinary perspective on hypothesis creation and validation using LLMs and related AI systems. Unlike previous works that focus narrowly on specific domains such as biomedicine \cite{sybrandt2017moliere, jumper2021highly} or materials science \cite{jain2020materials, gruver2024fine}, this survey highlights the versatility of LLMs across a diverse array of fields, including social sciences \cite{touvron2023llama}, environmental studies \cite{wang2023scimon}, and computational sciences \cite{fok2024search}. By synthesizing insights from state-of-the-art studies and frameworks, this survey bridges the gap between theoretical advancements and practical applications \cite{zhang2024comprehensive, lu2024ai}. This comprehensive approach not only underscores the transformative potential of LLMs but also illuminates their adaptability to the multifaceted challenges of contemporary scientific research.

In addition to delineating the opportunities presented by LLMs, this work systematically identifies critical barriers to their effective utilization. Key challenges include enhancing the novelty of generated hypotheses, improving the feasibility of proposed ideas, mitigating data biases, and addressing the interdisciplinary adaptability of AI-driven methodologies. To overcome these obstacles, this survey proposes actionable strategies such as the development of generative exploration models \cite{chen2024chemist}, the incorporation of human-in-the-loop systems \cite{tari2010discovering}, and the fine-tuning of models for domain-specific applications. Furthermore, by emphasizing principles of transparency, fairness, and inclusivity, this survey addresses the ethical and practical considerations associated with deploying LLMs in scientific discovery \cite{bommasani2021opportunities, shavit2023practices, fok2024search}. Through these contributions, this work provides a roadmap for harnessing the full potential of LLMs in advancing the frontiers of knowledge.

The remainder of this section outlines the prevailing approaches to scientific knowledge creation and validation, critiques their inherent limitations, and elaborates on the key contributions of this survey paper. In doing so, it establishes a foundation for understanding the transformative impact of LLMs and related AI systems on the landscape of scientific discovery.

\subsection{Overview of Current Approaches to Scientific Hypothesis Generation and Validation}

The integration of LLMs into scientific research has initiated a paradigm shift in how hypotheses are generated and validated. Leveraging their capacity to process and synthesize vast volumes of domain-specific data, LLMs empower researchers to uncover latent patterns, generate insights, and explore relationships that are often inaccessible through traditional methodologies. Their scalability and computational efficiency make LLMs well-suited to address complex, interdisciplinary challenges across diverse scientific domains~\cite{zhang2024comprehensive, lu2024ai}.
LLMs have significantly transformed the hypothesis generation process by enabling data-driven exploration across structured and unstructured sources. Knowledge graph-based systems, such as MOLIERE~\cite{sybrandt2018large, sybrandt2017moliere} and SciAgents~\cite{ghafarollahi2024sciagents}, facilitate the discovery of novel connections by mapping semantic relationships in fields like biomedicine and materials science. Complementing these, retrieval-augmented generation (RAG) frameworks, exemplified by VELMA~\cite{schumann2024velma} and Chemist-X~\cite{chen2024chemist}, integrate curated knowledge bases with generative modeling to produce hypotheses that are both contextually grounded and creatively extended.

These methodologies are complemented by multi-omics integration platforms, such as VirtualPlant~\cite{katari2010virtualplant} and BioLunar~\cite{wysocki2024llm}, which synthesize genomic and pharmacological data to foster cross-disciplinary discoveries. In materials science, AI-driven hypothesis generation has been instrumental in predicting new material properties~\cite{gruver2024fine}. Advancing hypothesis generation, machine learning models equipped with reinforcement learning capabilities adapt dynamically to data-rich, iterative environments, as is commonly observed in drug discovery~\cite{blanco2023role, tari2010discovering}. Text mining tools, like Dyport~\cite{tyagin2024dyport}, track the evolution of concepts across large textual datasets, enabling hypothesis creation in genomics and biomedical research. Other approaches, such as knowledge graph embeddings~\cite{wang2023scimon}, improve the scalability of hypothesis generation by structuring scientific knowledge into machine-readable formats. Together, these diverse methodologies illustrate how LLMs and AI-driven systems are reshaping the scientific discovery process, providing new pathways for interdisciplinary research and novel hypothesis generation. 

Hypothesis validation is equally important to the scientific process, with LLMs playing a critical role in ensuring that proposed ideas are both scientifically plausible and testable. Tools like Lab-bench~\cite{laurent2024lab} and AgentClinic~\cite{schmidgall2024agentclinic} facilitate experimental validation by simulating laboratory conditions to assess reproducibility and statistical significance. Simulation-based approaches provide cost-effective virtual environments for testing hypotheses in fields such as biomedicine~\cite{qi2024large, ghafarollahi2024protagents} and robotics~\cite{schumann2024velma}. Predictive model-based validation evaluates hypotheses using advanced metrics such as Bayesian posterior probability and prediction accuracy~\cite{jha2019hypothesis, chen2024chemist}. These methodologies are augmented by cross-domain validation systems, which test the generalizability of hypotheses across diverse scientific domains~\cite{sybrandt2017moliere, zhou2024hypothesisgenerationlargelanguage}. Additionally, crowdsourced validation platforms harness collective intelligence to evaluate hypotheses in social sciences and education, utilizing metrics like consensus scores~\cite{kim2018research, touvron2023llama}.

Multi-agent validation systems~\cite{ma2024sciagent} capitalize on distributed expertise to enhance collaborative validation processes, while causal inference frameworks~\cite{jha2019hypothesis, fok2024search} ensure the robustness of causal relationships through advanced structural causal modeling techniques. Dynamic validation tools~\cite{wang2023scimon, tang2024prioritizing} incorporate real-time data to continuously refine validation processes, offering metrics like anomaly detection precision. Benchmarking platforms~\cite{qi2024large, aubin2024llms, blanco2023role} ensure the reproducibility of validation efforts by measuring performance against standardized benchmarks. Moreover, iterative human-AI collaboration systems~\cite{shavit2023practices, tari2010discovering} combine human insights with AI-driven validation methodologies, enhancing explainability and user satisfaction.

The tools and datasets supporting these advancements form the foundation of contemporary knowledge creation and validation processes. Biomedical repositories like PubMed~\cite{pubmed}, Gene Ontology (GO)~\cite{ashburner2000gene}, and UK Biobank~\cite{ukbiobank} provide structured, high-quality data essential for hypothesis generation in areas such as genomics and drug discovery. Tools like MOLIERE~\cite{sybrandt2017moliere} and Chemist-X~\cite{chen2024chemist} effectively utilize these datasets to uncover novel connections by leveraging knowledge graphs and retrieval-augmented generation. Similarly, material science repositories, including the Materials Project~\cite{jain2020materials} and ChemBench~\cite{walker2010chembench}, offer comprehensive data on chemical compositions and properties, empowering researchers to hypothesize about new materials. AI-driven materials discovery frameworks, such as reinforcement learning-based methods~\cite{gruver2024fine}, further enhance the predictive accuracy of material property estimations. Interdisciplinary repositories, hosted by platforms like Hugging Face\footnote{https://huggingface.co/}, integrate multi-modal data to enable cross-domain hypothesis generation.

Simulation tools facilitate virtual testing environments, providing scalable validation methods for drug interactions, materials discovery, and robotics~\cite{qi2024large, schumann2024velma}. Meanwhile, open-source platforms like Hugging Face provide APIs and pre-trained models to streamline the integration of LLM capabilities into research workflows~\cite{touvron2023llama}. These platforms are increasingly adopted in both academic and industrial research, demonstrating their versatility in supporting large-scale scientific discovery. Through the synergistic application of these tools and methodologies, LLMs continue to advance scientific exploration by enabling more efficient and innovative approaches to hypothesis generation and validation. By grounding these processes in robust datasets and cutting-edge computational frameworks, researchers are well-positioned to uncover transformative insights that address some of the most pressing challenges of our time.

\subsection{Challenges in Scientific Hypothesis Generation and Validation}

Hypothesis creation, while significantly advanced by the capabilities of LLMs, faces several inherent limitations that impede the development of novel and impactful ideas. One critical challenge lies in the training paradigms of LLMs, which often replicate established knowledge patterns, thereby limiting the generation of truly innovative hypotheses. Techniques such as counterfactual reasoning and anomaly detection~\cite{fok2024search, wang2023scimon, weidinger2021ethical} are proposed to tackle the challenge. These techniques can be integrated into training processes to encourage deviation from conventional norms. Promoting novelty further requires the application of contrastive learning~\cite{hu2024comprehensive} and dynamic retraining~\cite{han1994dynamic} to explore uncharted scientific territories, especially in interdisciplinary contexts where cross-domain interactions hold the potential for groundbreaking insights~\cite{zhou2024hypothesisgenerationlargelanguage, chen2024chemist}.
However, achieving these insights requires using curated datasets and enhanced semantic mapping techniques to bridge disparate fields effectively~\cite{shavit2023practices, jha2019hypothesis}. The integration of structured knowledge graphs has been proposed to improve scientific hypothesis mapping by identifying non-obvious connections across domains~\cite{sybrandt2017moliere}. Additionally, the scalability of data integration remains a persistent issue, with the need for scalable architectures and real-time synthesis to manage large and diverse datasets~\cite{touvron2023llama, tang2024prioritizing}. AI-driven hypothesis generation frameworks must also address limitations related to model adaptability, ensuring that evolving knowledge bases are incorporated into the reasoning process~\cite{gruver2024fine}.
Another pressing concern is the interpretability of hypothesis generation systems, which often function as opaque ``black boxes''. Addressing this issue through logic-based reasoning and explainable AI (XAI) methodologies can significantly enhance trust and usability among researchers~\cite{shavit2023practices, tari2010discovering}. By integrating transparency-enhancing techniques such as causal inference models~\cite{jha2019hypothesis} and retrieval-augmented reasoning~\cite{fawzi2022discovering}, LLMs can improve their ability to justify generated hypotheses in a scientifically rigorous manner.

Equally critical in hypothesis development, the validation phase ensures that proposed ideas are scientifically plausible and relevant. However, LLMs often struggle with domain-specific adaptability, as nuanced validation criteria vary significantly across specialized fields. Modular architectures incorporating domain-specific constraints offer a promising solution for improving validation accuracy~\cite{jha2019hypothesis, tang2024prioritizing}. Another challenge is interdisciplinary validation, which requires adaptive systems capable of activating discipline-specific submodules to ensure the relevance of hypotheses across various fields~\cite{zhou2024hypothesisgenerationlargelanguage, sybrandt2017moliere}.
Furthermore, the reliance on computational metrics for validation often disconnects hypotheses from real-world feasibility. Simulation tools provide scalable virtual environments that can pre-test hypotheses effectively~\cite{qi2024large, schumann2024velma}. These methods have proven especially useful in biomedical research and materials science, where experimental verification can be resource-intensive~\cite{gruver2024fine}. However, high-risk, potentially transformative hypotheses are often penalized by conventional validation metrics that favor incremental advancements. To address this, risk-weighted evaluation frameworks must balance potential impact with empirical reliability~\cite{shavit2023practices, fok2024search}.
Additionally, multi-criteria validation approaches, combining metrics such as relevance, novelty, and feasibility, can provide a holistic assessment of hypothesis quality~\cite{qi2024large, aubin2024llms, 10.1145/3613905.3636301}. Incorporating human-in-the-loop validation frameworks can further enhance hypothesis evaluation by integrating expert feedback into AI-driven validation systems~\cite{tari2010discovering, wang2023scimon}. These methodologies contribute to a more rigorous and adaptable validation process, ensuring that LLM-generated hypotheses align with real-world applicability and scientific standards.

\textbf{Limitations in Achieving Novelty and Feasibility with LLMs.} The pursuit of novelty and feasibility in hypothesis creation is central to scientific progress, yet LLMs face fundamental challenges in both areas. A key issue lies in their training paradigm: LLMs are primarily trained on large corpora of existing, historically grounded data, which inherently biases them toward established knowledge patterns rather than fostering the generation of disruptive or paradigm-shifting insights \cite{zhou2024hypothesisgenerationlargelanguage, ghafarollahi2024sciagents}. This reliance results in a regression to the mean in idea generation, where LLMs tend to prioritize statistically likely continuations over epistemic risk-taking, even when prompted to be creative. Consequently, LLMs often generate variants of well-known ideas and rarely propose counterfactuals or unconventional hypotheses \cite{fok2024search, jha2019hypothesis}.

To overcome this limitation, several techniques have been proposed. Contrastive learning and generative exploration models have been developed to encourage semantic divergence and novelty \cite{wang2023scimon, fawzi2022discovering, tang2024prioritizing}. Reinforcement learning with novelty-seeking reward signals has also shown promise in promoting the exploration of low probability, high impact hypotheses \cite{gruver2024fine, blanco2023role}. However, these approaches remain experimental and require careful design to avoid compromising scientific rigor. In addition, novelty thresholds vary across disciplines, necessitating dynamic adjustment mechanisms within LLMs to align with domain-specific standards \cite{zhou2024hypothesisgenerationlargelanguage, shavit2023practices}.
Generating cross-disciplinary insights poses further complexity. Effective novelty requires semantic mapping across curated, multi-disciplinary datasets to uncover novel associations \cite{ghafarollahi2024sciagents, sybrandt2017moliere}. Yet, data bias and conservatism remain persistent obstacles. Models trained on historical data often struggle to produce original ideas, reinforcing conventional thinking. Diversified training corpora and novelty-boosting algorithms have been proposed to mitigate these effects \cite{chen2024chemist, wang2023scimon}. Additionally, surface-level similarity metrics are insufficient for detecting deeper conceptual innovations, underscoring the need for models capable of identifying unique theoretical implications \cite{gruver2024fine}. Retrieval augmented reasoning can enhance this capability by grounding hypotheses in diverse, relevant contexts \cite{jha2019hypothesis}.

Feasibility presents a parallel set of challenges. LLMs often generate hypotheses without considering practical constraints, limiting their real-world applicability. Effective feasibility requires grounding in empirical evidence and domain-specific constraints. Techniques such as the integration of multi-modal data, including experimental results and sensor outputs, have been proposed to improve feasibility assessment \cite{qi2024large, ghafarollahi2024sciagents}. Foundation models with physical interaction capabilities can further bridge the gap between theory and experimental validation \cite{blanco2023role, jha2019hypothesis}. For example, models integrated with robotic platforms, such as those used in automated laboratories or robotic chemists~\cite{king2011rise, fakhruldeen2022archemist}, can physically conduct experiments based on model-generated hypotheses, enabling real-time feedback and refinement. These systems, like the "robot scientist" Eve~\cite{williams2015eve} used in drug discovery, embody physical interaction by selecting compounds, operating lab equipment, and analyzing results, closing the loop between hypothesis generation and empirical testing.
In data-scarce scientific domains, synthetic data generation and few-shot learning can supplement existing datasets to improve generalization \cite{wang2023scimon, chen2024chemist}. Embedding field-specific constraints into LLM pipelines, such as resource availability or laboratory feasibility, can ensure that generated hypotheses are actionable \cite{gruver2024fine, shavit2023practices}. Simulation environments like ChemBench provide scalable virtual platforms for early-stage hypothesis testing \cite{walker2010chembench}. In biomedical and materials science, digital twin simulations offer higher fidelity feasibility assessments by emulating real-world experimental settings \cite{fawzi2022discovering, tang2024prioritizing}. Tailoring feasibility metrics to reflect domain-specific requirements remains essential for producing hypotheses that are both practical and impactful \cite{aubin2024llms, touvron2023llama}.

\textbf{Ethical Concerns in LLM-Driven Hypothesis Generation and Validation.} As LLMs gain autonomy in scientific research, several ethical challenges emerge that impact the trustworthiness, accountability, and inclusivity of AI-assisted discovery. First, LLMs trained on biased data may reproduce and even amplify social, demographic, and epistemic inequities, potentially marginalizing underrepresented perspectives or reinforcing dominant paradigms~\cite{shavit2023practices, weidinger2021ethical}. This risk is particularly serious in fields like biomedicine and public policy, where disparities in training data can lead to flawed or discriminatory hypotheses.
Second, the phenomenon of AI hallucination~\cite{venkit2024audit} presents a threat to scientific integrity. LLMs may generate outputs that appear fluent and scientifically plausible but are factually incorrect or unsupported by evidence. Without transparent reasoning or source attribution, these outputs can mislead researchers and corrupt the hypothesis evaluation pipeline~\cite{tang2024prioritizing}. This is especially dangerous when outputs are integrated into high-stakes domains such as clinical decision-making or environmental policy.
Third, accountability is difficult to establish when LLM-generated hypotheses lead to errors or unintended consequences. Whether responsibility should lie with model developers, research users, or deploying institutions remains unclear. In the absence of mechanisms for provenance tracking, version control, and responsible usage documentation, resolving these questions remains difficult~\cite{jaradeh2019open}.
Ethical safeguards must be embedded throughout the hypothesis generation and validation pipeline to mitigate these concerns. These include explainable AI (XAI) methods, allowing researchers to understand and verify model reasoning, and human-in-the-loop frameworks integrating expert oversight into model output evaluation~\cite{jaradeh2019open, tang2024prioritizing}. Auditing protocols and model cards~\cite{kennedy2025model, nunes2024using} can also help disclose ethical risks, intended use cases, and model limitations, promoting transparency and accountability.

\textbf{Regulatory and Policy Implications.} The increasing integration of LLMs into scientific workflows presents a growing governance challenge. Despite their widespread use, few regulatory standards exist for evaluating or auditing AI-generated hypotheses. Traditional scientific metrics do not adequately capture emerging concerns introduced by LLMs, such as ethical risk, reproducibility, and explainability. To ensure responsible deployment, domain-specific policy frameworks are needed to support traceable, verifiable, and equitable use of LLMs in science. These frameworks should include auditing tools to verify provenance, model documentation standards, and transparent disclosures of AI contributions in research publications. Open science principles must be preserved by promoting access to open-source models and datasets, especially for researchers in low-resource environments. Institutions, funding agencies, and regulatory bodies should adapt review criteria and funding guidelines to account for the role of LLMs in the research process. This includes evaluating whether proper oversight, validation mechanisms and ethical safeguards are in place. Without these changes, scientific discovery risks becoming dependent on opaque systems that operate without clear accountability or alignment with community norms.

\subsection{Contributions of this Study}

This survey paper comprehensively examines the challenges and opportunities in automating hypothesis creation and validation using LLMs. As LLMs increasingly contribute to scientific research across diverse disciplines, understanding their limitations and potential is critical for fostering innovation. Despite their transformative capabilities, LLMs face significant challenges in generating novel, feasible, and impactful hypotheses. By consolidating methodologies, identifying gaps, and proposing actionable strategies, this survey equips the research community with a roadmap to enhance the effectiveness of LLMs in hypothesis generation and validation.

The primary contributions of this study are as follows:
\begin{itemize}
    \item \textbf{Comprehensive Review of Current Approaches:} This survey offers a structured, interdisciplinary overview of current approaches to hypothesis creation and validation. It synthesizes insights from diverse scientific fields, bridging gaps in the literature and providing a unified perspective to guide future research directions.
    \item \textbf{Identification of Challenges and Gaps:} Key challenges in hypothesis creation and validation, such as promoting novelty, improving feasibility assessments, and addressing data biases, are systematically analyzed. These insights illuminate where LLMs fall short and provide clarity for researchers and practitioners aiming to develop robust and innovative systems.
    \item \textbf{Interdisciplinary Relevance:} Highlighting the versatility of LLMs, this survey demonstrates their applicability across fields such as biomedicine, materials science, social sciences, and environmental research. This survey illustrates how LLMs can adapt to varied scientific contexts and foster innovation across disciplines by showcasing use cases and domain-specific challenges.
    \item \textbf{Establishing a Framework for Ethical and Practical Applications:} Ethical and practical considerations are emphasized to ensure that LLMs are deployed responsibly and effectively. This survey sets foundational guidelines for creating transparent, fair, and inclusive systems, fostering trust and usability in scientific workflows.
    \item \textbf{Actionable Insights for Transformative Advancements:} By encouraging a shift from static, one-size-fits-all approaches to dynamic, adaptive, and interdisciplinary systems, this survey aligns with the evolving demands of scientific research. It highlights how data-driven methodologies can redefine the pace and scope of discovery, providing tools and insights for transformative advancements in hypothesis creation and validation.
    \item \textbf{Proposal of Novel Strategies and Future Directions:} This survey presents forward-looking strategies to overcome identified challenges. These include generative exploration models, hybrid human-AI systems, and risk-tolerant validation frameworks that equip researchers and developers with practical tools to enhance LLM-driven scientific exploration.
    
\end{itemize}

In summary, this survey addresses the critical need for a holistic, interdisciplinary resource that integrates theoretical advancements with practical applications. By identifying gaps, offering actionable strategies, and emphasizing ethical considerations, to empower researchers, practitioners, and policymakers to harness the transformative potential of LLMs for scientific innovation. This survey offers a foundation for exploring approaches to hypothesis creation and validation, contributing to the development of data-driven scientific discovery.
\section{Definitions and Overview}

This section introduces key concepts essential to this survey and outlines the structure of the paper, providing a roadmap for the topics discussed in subsequent sections.

\subsection{Definitions}

\noindent \textbf{Hypothesis:} A hypothesis is a tentative explanation, relationship, or proposition that can be empirically tested or theoretically evaluated. In scientific research, hypotheses serve as foundational units for exploration, guiding the formulation of experiments or analytical tasks. Within the context of LLM-driven scientific discovery, a hypothesis can range from a declarative statement proposing a causal relationship (e.g., "Gene X inhibits Protein Y") to an abstract proposition extracted or synthesized from unstructured data (e.g., text-based summaries or concept clusters).

\noindent \textbf{Novelty:} Novelty in hypothesis creation~\cite{witt2009propositions, hallsworth2023scientific} refers to the degree to which a generated hypothesis differs from existing knowledge. It indicates the originality of the hypothesis by measuring its divergence from known concepts, relationships, or patterns in a given domain. Quantifying novelty is essential for evaluating whether hypotheses contribute meaningful advancements rather than reiterating existing ideas~\cite{zhou2024hypothesisgenerationlargelanguage, fok2024search, shibayama2021measuring}.  

Novelty can be quantified using similarity or distance metrics. Given a generated hypothesis $H$ represented as a vector $h$ in an embedding space, and a set of existing hypotheses $H_i$ with vectors $h_i$, the novelty $N(H)$ of $H$ can be computed as the inverse of the average cosine similarity between $h$ and each $h_i$:  
\begin{align}
    N(H) = 1 - \frac{1}{|S|} \sum_{i \in S} \text{cosine\_similarity}(h, h_i),
\end{align}  
where $S$ is the set of existing hypotheses.  

High novelty implies \textit{low similarity} to existing knowledge, suggesting \textit{originality} and a potential breakthrough or unexplored idea. However, ensuring novelty while maintaining scientific validity is a challenging balance, as excessive deviation from known knowledge can lead to implausible or untestable hypotheses~\cite{gruver2024fine, shavit2023practices}. Contrastive learning and retrieval-augmented reasoning have been proposed as strategies to improve novelty in LLM-generated hypotheses by refining semantic divergence while preserving logical consistency~\cite{ kim2024enhancingcontrastivelearningefficient}.

\noindent \textbf{Feasibility:} Feasibility~\cite{walker1987feasible, song2024step} assesses whether a generated hypothesis is practically testable and grounded within the known constraints of the domain. It evaluates the likelihood that a hypothesis can be experimentally validated or implemented in real-world settings. Feasibility is particularly critical in domains such as biomedicine and materials science, where empirical validation requires substantial experimental resources~\cite{wang2024survey}.  
The feasibility $F(H)$ of a hypothesis $H$ can be calculated as a weighted combination of empirical and theoretical scores. If $f_{empirical}$ represents empirical feasibility, for instance, the availability of necessary data, equipment, or methods, and $f_{theoretical}$ represents the theoretical validity based on domain knowledge, the feasibility is defined as:  
\begin{align}
F(H) = w_{emp} \cdot f_{empirical} + w_{theo} \cdot f_{theoretical},
\end{align}
where $w_{emp}$ and $w_{theo}$ are weights reflecting the importance of each factor, subject to $ w_{emp} + w_{theo} = 1 $. A high feasibility score indicates that the hypothesis is both theoretically sound and practically achievable, increasing its potential value for real-world testing~\cite{shavit2023practices, jha2019hypothesis}.  

Multi-modal feasibility assessments, incorporating experimental data, sensor inputs, and real-world simulations, have been proposed to improve LLM-driven hypothesis validation~\cite{fawzi2022discovering, chen2024chemist}. AI-powered frameworks such as digital twin simulations~\cite{tang2024prioritizing} and human-in-the-loop validation further enhance feasibility by dynamically refining hypotheses based on real-time feedback. Tailoring feasibility metrics to discipline-specific challenges ensures that LLM-generated hypotheses remain actionable and impactful in scientific research~\cite{aubin2024llms, touvron2023llama}.

\noindent \textbf{Open Domain:}  
An open domain in hypothesis generation refers to a broad, unrestricted field where generated hypotheses are not confined to specific topics or predefined structures. In open-domain settings, hypotheses may span across multiple fields, incorporating interdisciplinary knowledge~\cite{zhou2024hypothesisgenerationlargelanguage, ghafarollahi2024sciagents}.  
The open domain can be represented as a large, unrestricted hypothesis space $\mathcal{H}_{open}$, where any hypothesis $H$ is possible. It is formally defined as:  
\begin{align}
\mathcal{H}_{open} = \{H \ | \ H \in \text{Any Topic or Field} \}.
\end{align}  
Open-domain hypothesis generation requires models that can generalize across diverse fields, enabling novel, interdisciplinary insights, but often making validation and relevance assessment more challenging~\cite{sybrandt2018large, wang2023scimon}. Techniques such as retrieval-augmented generation (RAG) and knowledge graph-based reasoning have been employed to structure open-domain hypothesis generation, ensuring hypotheses remain scientifically plausible while preserving creativity~\cite{chen2024chemist, jha2019hypothesis}.  
One of the key challenges in open-domain hypothesis generation is contextual grounding, as LLMs may generate syntactically correct hypotheses but lack empirical feasibility~\cite{shavit2023practices, aubin2024llms}. To address this, adaptive fine-tuning methods that incorporate domain constraints dynamically have been proposed to improve the relevance of generated hypotheses while maintaining the breadth of interdisciplinary exploration~\cite{tang2024prioritizing, touvron2023llama}. 

\noindent \textbf{Closed Domain:}  
A closed domain refers to a specific, restricted area of knowledge in which generated hypotheses are limited to a particular subject or closely related subfields. In a closed-domain setting, hypotheses are generated with narrowly defined constraints, making them more easily verifiable within the domain~\cite{qi2024large, gruver2024fine}.  
Closed domain can be represented by a bounded hypothesis space $ \mathcal{H}_{closed} $, restricted by a set of domain constraints $C$:  
\begin{align}
\mathcal{H}_{closed} = \{H \ | \ H \text{ adheres to constraints } C \},
\end{align}  
where $C$ includes specific requirements or parameters related to the domain (e.g., biomedicine, materials science). Closed-domain hypothesis generation enhances relevance and ease of validation but may limit the scope for cross-disciplinary or radically innovative insights~\cite{jha2019hypothesis, shavit2023practices}.  
To improve domain-specific accuracy, retrieval-augmented generation (RAG) and fine-tuned foundation models have been employed in closed-domain settings~\cite{chen2024chemist, wang2023scimon}. Tools such as MOLIERE for biomedical hypothesis validation~\cite{sybrandt2018large} and SciAgents for structured scientific discovery~\cite{ghafarollahi2024sciagents} demonstrate the efficacy of constrained knowledge generation. However, over-restriction in closed-domain settings may reduce the exploratory potential of LLMs, necessitating hybrid approaches that allow limited cross-domain adaptation while maintaining domain constraints~\cite{tang2024prioritizing, touvron2023llama}.  

\noindent \textbf{Relevance of Generated Hypothesis:} Relevance assesses how well a hypothesis aligns with the target scientific community's goals, needs, or pressing questions. It measures contextual importance within the domain~\cite{wang2023scimon}. Relevance can be computed via citation-based retrieval scores, topic overlap, or alignment with funding priorities.

\noindent \textbf{Quality of Generated Hypothesis:}  
The quality of a generated hypothesis reflects its potential scientific impact, evaluated through a combination of novelty, feasibility, and relevance. A high-quality hypothesis introduces new achievable and significant insights within the scientific domain.  
The quality $Q(H)$ of a hypothesis $H$ can be expressed as a weighted aggregate of novelty $N(H)$, feasibility $F(H)$, and relevance $R(H)$:  
\begin{align}
Q(H) = w_N \cdot N(H) + w_F \cdot F(H) + w_R \cdot R(H),
\end{align}  
where $w_N$, $w_F$, and $w_R$ are weights assigned based on the importance of each component for the specific research objective, and $w_N + w_F + w_R = 1$. A hypothesis with high-quality scores well across novelty, feasibility, and relevance, making it a strong candidate for empirical testing and potential scientific advancement~\cite{wang2023scimon, jha2019hypothesis}.  
Multi-criteria evaluation frameworks have been proposed to improve hypothesis quality assessment, integrating both quantitative and qualitative metrics~\cite{shavit2023practices, chen2024chemist}. For instance, contrastive learning techniques enable LLMs to refine hypotheses by balancing novelty and plausibility, reducing the risk of generating implausible or irrelevant ideas~\cite{hu2024comprehensive}. Similarly, human-in-the-loop validation allows domain experts to iteratively refine LLM-generated hypotheses, ensuring that AI-driven scientific discoveries align with empirical research priorities~\cite{aubin2024llms, touvron2023llama}.

\noindent \textbf{Bad Hypothesis:}  
A bad hypothesis is one that is either trivial, incorrect, non-novel, or infeasible. Such a hypothesis lacks scientific merit due to redundancy with existing knowledge, conceptual errors, or impracticality for empirical testing. The criteria for categorizing a generated hypothesis as bad include:  

\begin{enumerate}
    \item \textit{Low Novelty:} Highly similar to known hypotheses, offering little or no new information~\cite{qi2024large}.
    \item \textit{Low Feasibility:} Practically or theoretically untestable, making empirical validation unrealistic~\cite{kim2018research}.
    \item \textit{Conceptually Incorrect:} These hypotheses violate domain logic or introduce contradictions. They are implicitly penalized via low $f_{theoretical}$ in feasibility and can also be flagged using symbolic consistency checks or NLI-based verification frameworks~\cite{zhou2024hypothesisgenerationlargelanguage, sybrandt2018large}.
\end{enumerate}  

Let $N(H)$ denote novelty, $F(H)$ feasibility, and $E(H)$ represent correctness (with $E(H) = 1$ for a valid hypothesis and $0$ for a conceptually flawed hypothesis). A hypothesis $H$ is classified as bad if:  
\begin{align}
N(H) < \tau_N \quad \text{or} \quad F(H) < \tau_F \quad \text{or} \quad E(H) = 0,
\end{align}  
where $\tau_N$ and $\tau_F$ are thresholds for novelty and feasibility, respectively. These thresholds are domain-specific. For instance, earlier studies~\cite{sybrandt2018large, tang2024prioritizing} employ percentile-based thresholds derived from baseline distributions, whereas others adopt tunable cutoffs informed by downstream validation outcomes or expert evaluations. Identifying and filtering bad hypotheses can prevent resource wastage in experimental stages, allowing focus on hypotheses with potential scientific value~\cite{beltagy2019scibert}.  

LLMs can be enhanced through knowledge-driven hypothesis generation that leverages structured data sources and domain expertise to mitigate the generation of bad hypotheses. Additionally, retrieval-augmented validation frameworks such as SciFact and MOLIERE improve hypothesis filtering by cross-referencing scientific literature. AI-driven hypothesis verification techniques, integrating NLP-based consistency checks, further enhance the rejection of incorrect or redundant hypotheses.

\begin{figure*}
    \centering
    \includegraphics[width=\linewidth, height=1.2\linewidth]{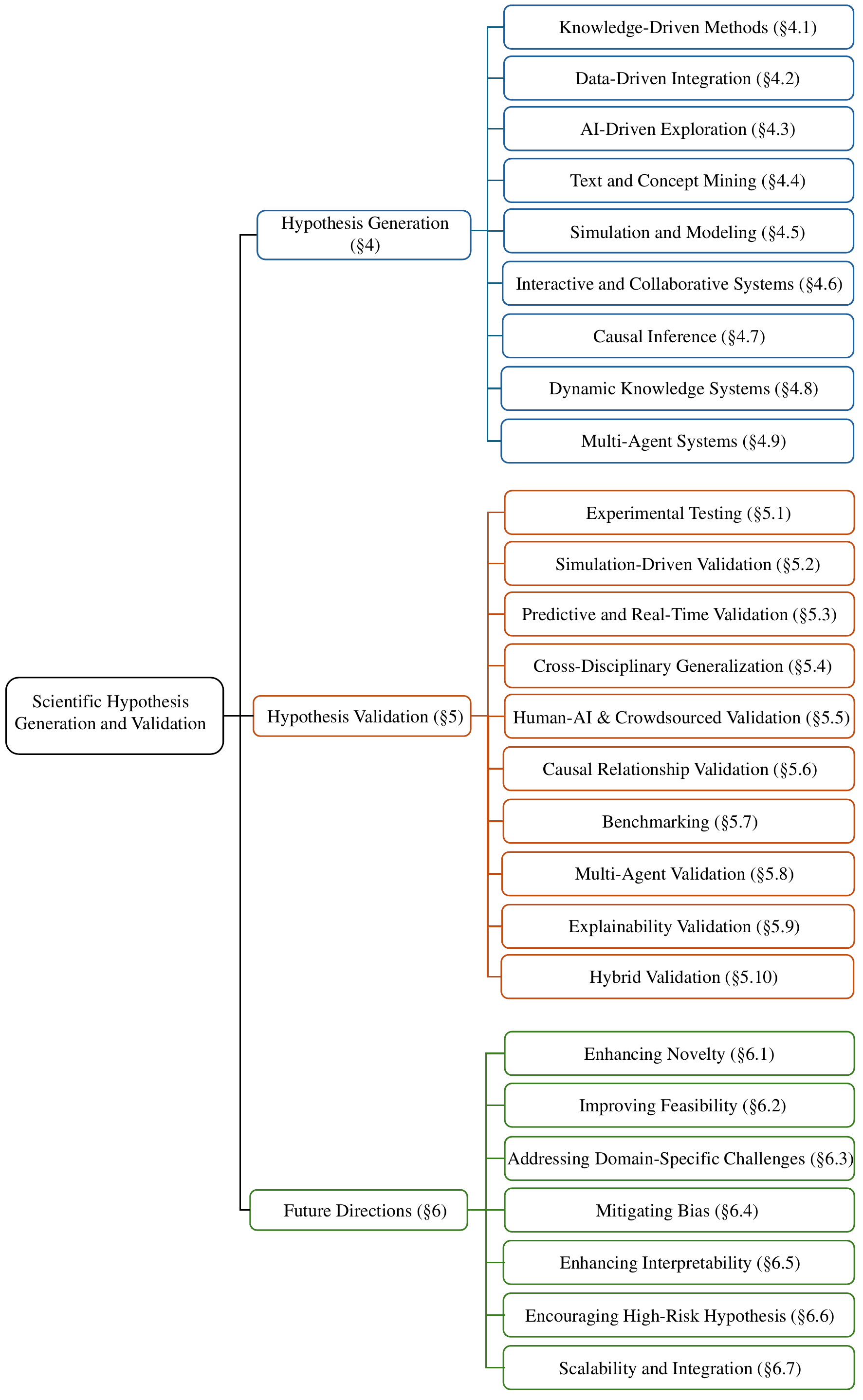}
    \caption{Flow diagram of the survey structure. This figure guides the reader through the organization of the paper, beginning with hypothesis creation approaches (§4), progressing through hypothesis validation methods (§5), and culminating in open challenges and future directions (§6). It highlights how various components of scientific discovery, from AI-driven exploration to validation frameworks, are interconnected in the survey's narrative.}
    \label{fig:figure_2}
\end{figure*}

\subsection{Overview and Structure of the Paper}

This survey provides a systematic and comprehensive exploration of scientific knowledge creation and validation using Large Language Models (LLMs). It is organized into sections that address the methodologies, challenges, and opportunities in advancing this field. The structure is designed to guide readers through an understanding of the state-of-the-art, the limitations of current systems, and the opportunities for future innovation. Figure~\ref{fig:figure_2} provides an overview of the survey structure.

\begin{itemize}

    \item \textbf{Datasets Supporting Hypothesis Generation and Validation (Section~\ref{sec:datasets}):} Data serves as the foundation for hypothesis generation and validation. Section~\ref{sec:datasets} describes available datasets (e.g., SciFact, PubMedQA, AVeriTeC) and tools like MOLIERE, ChemBench, and Project Jupyter that support hypothesis analysis and validation. It highlights the importance of understanding the role of datasets, their domain-specific applications, and the challenges of ensuring data quality, fairness, and accessibility to enable meaningful scientific exploration.
    
    \item \textbf{Hypothesis Generation Approaches (Section~\ref{sec:hypothesis_generation}):} Section~\ref{sec:hypothesis_generation} explores the diverse methodologies employed for hypothesis generation, emphasizing the need to understand how LLMs identify patterns, uncover relationships, and generate new ideas. It highlights the importance of categorizing approaches to assess their strengths and limitations in addressing various scientific challenges.

    \item \textbf{Hypothesis Validation Approaches (Section~\ref{sec:hypothesis_validation}):} Validation is a critical step in ensuring that hypotheses are scientifically plausible and impactful. Section~\ref{sec:hypothesis_validation} discusses the necessity of robust validation methods to evaluate hypotheses' feasibility, novelty, and relevance while addressing the constraints posed by different scientific contexts.

    \item \textbf{Future Directions for Advancing the Field (Section~\ref{sec:future_directions}):} Building on the identified challenges, Section~\ref{sec:future_directions} motivates the need for actionable strategies to enhance the capabilities of LLMs. It discusses the potential of integrating advanced methodologies and fostering interdisciplinary collaboration to push the boundaries of hypothesis creation and validation.

    \item \textbf{Conclusion (Section~\ref{sec:conclusion}):} The paper concludes in Section~\ref{sec:conclusion} by reflecting on the transformative potential of LLMs in scientific discovery. It emphasizes the urgency of adopting innovative approaches to maximize their impact and advocates for a shift toward systems that prioritize creativity, adaptability, and interdisciplinary applicability.
\end{itemize}
\section{Datasets}
\label{sec:datasets}

Datasets are the cornerstone of hypothesis creation and validation, providing the foundational information required to generate, test, and refine hypotheses. Their selection influences not only the novelty and feasibility of hypotheses but also their scope and applicability across domains. From biomedicine and materials science to social science and artificial intelligence, datasets reflect the diverse nature of scientific inquiry and drive advancements by enabling tailored hypothesis generation and rigorous validation.

This section reviews key datasets widely used across scientific fields, categorizing them by structure, domain, and functionality. By exploring curated knowledge graphs, textual corpora, and specialized scientific data repositories, we highlight their role in supporting innovation, addressing complex challenges, and fostering interdisciplinary research. The systematic analysis of these datasets provides insights into their characteristics, evaluation metrics, and potential for enabling groundbreaking discoveries.

\begin{table}[ht]
\centering
\caption{Summary of Datasets for Hypothesis Creation and Validation (Part 1)}
\resizebox{\columnwidth}{!}{%
\begin{tabular}{p{1.8cm}|p{3cm}|p{2.5cm}|p{1.5cm}|p{2.5cm}|p{1.6cm}|p{0.8cm}|p{0.8cm}}
\hline
\textbf{Dataset Name} & \textbf{Description} & \textbf{Statistics} & \textbf{Domain} & \textbf{Evaluation Metrics} & \textbf{Modality} & \textbf{Novelty (Y/N)} & \textbf{Feasibility (Y/N)} \\ \hline
PubMed Abstracts~\cite{pubmed} & Biomedical literature database for hypothesis generation in biology and medicine & Over 34 million abstracts & Biomedicine & Relevance, Co-occurrence Analysis, ROC-AUC & Text & Y & Y \\ \hline
MeSH~\cite{mesh} & Medical Subject Headings for categorizing PubMed content & 27,883 descriptors & Biomedicine & Similarity Metrics, Novelty Scoring & Text, Structured Metadata & Y & Y \\ \hline
ChEMBL~\cite{chembl} & Bioactive molecule database for drug discovery & Over 2 million compounds & Chemistry & Molecular Similarity, Drug-Likeness Scores & Text, Numerical & Y & Y \\ \hline
GENIA Corpus~\cite{kim2003genia} & Annotated biomedical text corpus for NLP tasks & Over 2,000 abstracts annotated with biomedical terms & Biomedicine & Precision, Recall, F1-Score & Text & N & N \\ \hline
Open Graph Benchmark (OGB)~\cite{hu2020open} & Large-scale graph datasets for hypothesis testing on network data & Over 100 datasets for benchmarking graph-based tasks & AI, Graph Analysis & Graph Accuracy, Link Prediction Accuracy & Graphs & Y & Y \\ \hline
UK Biobank~\cite{ukbiobank} & Longitudinal dataset linking genetic and phenotypic data & Data from over 500,000 participants & Genomics & Correlation, Causal Inference & Text, Numerical, Genomic & N & Y \\ \hline
MATBench~\cite{dunn2020benchmarking} & Material property prediction dataset for materials discovery & Includes over 100,000 material samples & Materials Science & Prediction Accuracy, RMSE & Numerical, Structured Data & Y & Y \\ \hline
ClimateNet~\cite{prabhat2020climatenet} & Dataset for climate change research and hypothesis validation & Over 10 years of climate observation data & \makecell[t]{Environ-\\mental\\ Science} & Temporal Trends, Anomaly Detection & Text, Numerical, Satellite Imagery & N & N \\ \hline
COCO Dataset~\cite{lin2014microsoft} & Dataset for image captioning and computer vision tasks & Over 300,000 images with captions & AI, Vision-Language Integration & Precision, BLEU Score & Image, Text & N & N \\ \hline
Gene Ontology (GO)~\cite{ashburner2000gene} & Hierarchical vocabulary for annotating gene products, supporting hypothesis generation and validation in genomics and molecular biology & Over 44,000 terms across three domains: biological process, molecular function, and cellular component & Genomics, Molecular Biology & Semantic Similarity, Graph-Based Validation, Annotation Consistency & Text, Graphs & N & Y \\ \hline
AHTech Electrolyte Additive Dataset~\cite{doi:10.1126/sciadv.adu4391}& High-throughput electrochemical screening data for electrolyte additives in aqueous zinc batteries & 180 candidates, 200-cycle efficiency per sample & \makecell[t]{Electro-\\chemistry,\\ Energy \\Storage} & Coulombic Efficiency, Additive Ranking, Feature Correlation & Tabular, Text & Y & Y \\ \hline
CSKG-600~\cite{BORREGO2025113280} & Expert-labeled hypothesis triples from a scholarly knowledge graph & 600 hypothesis statements with expert validation & Knowledge Graphs, Scholarly AI & Precision@K, Link Prediction Accuracy, Expert Agreement & Graphs, Text & Y & Y \\ \hline
\end{tabular}
}
\label{tab:datasets_summary_part1}
\end{table}

\begin{table}[ht]
\centering
\caption{Summary of Datasets for Hypothesis Creation and Validation (Part 2)}
\resizebox{\columnwidth}{!}{%
\begin{tabular}{p{2cm}|p{3cm}|p{2cm}|p{2cm}|p{2cm}|p{2.2cm}|p{0.8cm}|p{0.8cm}}
\hline
\textbf{Dataset Name} & \textbf{Description} & \textbf{Statistics} & \textbf{Domain} & \textbf{Evaluation Metrics} & \textbf{Modality} & \textbf{Novelty (Y/N)} & \textbf{Feasibility (Y/N)} \\ \hline
DrugBank~\cite{wishart2018drugbank} & Comprehensive dataset of drug-target interactions for drug discovery & Over 14,000 drugs and 6,000 protein targets & Pharmacology & Interaction Accuracy, Drug Efficacy Scores & Text, Structured Tables & Y & Y \\ \hline
AI2 Science Questions~\cite{clark2018think} & Dataset for hypothesis testing in question answering and reasoning & Over 10,000 multiple-choice science questions & Education, AI Reasoning & Accuracy, Explainability Metrics & Text, Structured QA Format & Y & N \\ \hline
Materials Project~\cite{jain2013commentary} & Database of materials properties and structures & Over 133,500 materials & Materials Science & Structural Matching, Novelty Filtering & Numerical, Structured Data & Y & Y \\ \hline
KEGG Pathway~\cite{kanehisa2000kegg} & Database of metabolic and signaling pathways & 540 pathways across species & Biomedicine, Genomics & Pathway Enrichment, Graph Metrics & Text, Graphs & Y & Y \\ \hline
American Community Survey (ACS)~\cite{acs} & Annual survey capturing demographic and social data in the US & 2.5 million responses/year & Social Science & Statistical Analysis, Demographic Comparisons & Text, Numerical & N & Y \\ \hline
Patent Data (USPTO)~\cite{uspto} & Text and metadata of granted patents & Over 10 million patents & Technology, Innovation & Citation Analysis, Novelty Score & Text, Structured Metadata & Y & N \\ \hline
XSum~\cite{narayan2018don} & Summarization dataset with diverse topics for NLP research & 226,711 summaries & NLP/Text Mining & BLEU, ROUGE for validation & Text & Y & Y \\ \hline
Cosmic~\cite{bamford2004cosmic} & Somatic mutation data for cancer research & 30,000 genes, 2 million mutations & Cancer Genomics & Mutation Analysis, Pathway Mapping & Text, Numerical & N & Y \\ \hline
Open Research Knowledge Graph (ORKG)~\cite{jaradeh2019open} & Structured knowledge graph of research contributions & 3 million triples & Multidisciplinary & Graph Centrality, Novelty Detection & Text, Graphs, Structured Metadata & Y & Y \\ \hline
\end{tabular}
}
\label{tab:datasets_summary_part2}
\end{table}

\textbf{PubMed~\cite{pubmed}:} PubMed is a foundational dataset in biomedical research, featuring over 34 million abstracts and citations across clinical medicine, pharmacology, and molecular biology. It supports hypothesis creation and validation by enabling co-occurrence analysis, uncovering relationships between biomedical entities, and facilitating natural language processing tasks such as entity recognition and document classification. Key evaluation metrics include co-occurrence analysis for identifying term relationships, ROC-AUC for assessing model performance in pattern recognition, and relevance metrics for aligning abstracts with hypotheses. These tools ensure robust hypothesis testing and data-driven insights. PubMed fosters novelty by revealing previously unlinked relationships while providing evidence-based validation through its peer-reviewed content, making it indispensable for biomedical research.

\textbf{MeSH (Medical Subject Headings)~\cite{mesh}:} MeSH is a structured controlled vocabulary that categorizes biomedical content in databases such as PubMed, providing a hierarchical framework for precise hypothesis generation and validation. It facilitates the categorization of biomedical content, linking terms to established descriptors to support hypothesis validation and enabling similarity and novelty scoring for hypothesis evaluation. Key metrics include semantic similarity, which quantifies relationships using ontological structures, novelty scoring to measure the uniqueness of hypotheses, and term coverage to assess alignment with MeSH categories. MeSH supports novel hypothesis creation by enabling semantic exploration and unique term connections, while its hierarchical structure ensures reliability in hypothesis validation.

\textbf{ChEMBL~\cite{chembl}:} ChEMBL is a comprehensive database of bioactive molecules designed for hypothesis generation and validation in drug discovery and medicinal chemistry. It facilitates the exploration of drug-target interactions, validates predictive models for structure-activity relationships (SAR), and enables analysis of compound efficacy and bioactivity. The dataset includes over 2 million molecules with bioactivity data, covering more than 14,000 biological targets. Metrics such as molecular similarity, drug-likeness scores, and binding affinity prediction accuracy support novel drug candidates and validate bioactivity insights. ChEMBL’s experimental data ensures both novelty and feasibility in hypothesis testing.

\textbf{GENIA Corpus~\cite{kim2003genia}:} The GENIA Corpus is an annotated biomedical text dataset tailored for NLP tasks, facilitating hypothesis generation and validation in text mining and entity recognition. Comprising over 2,000 abstracts annotated with more than 36,000 unique terms, it focuses on biomedical entities like proteins and genes, emphasizing transcription factors and cellular signaling. Its primary use is benchmarking NLP models, with evaluation metrics such as precision, recall, and F1-score for entity recognition. While it excels in validating NLP techniques, its primary role is not novel hypothesis generation.

\textbf{Open Graph Benchmark (OGB)~\cite{hu2020open}:} The Open Graph Benchmark (OGB) is a collection of over 100 graph datasets across domains like biology, chemistry, and computer science, designed for benchmarking machine learning models in graph-based tasks. It supports hypothesis generation about structural patterns in networks and validates graph-based hypotheses using metrics like graph accuracy, link prediction accuracy, and clustering coefficients. OGB provides predefined training, validation, and testing splits, making it ideal for exploring novel structural patterns and systematically validating hypotheses.

\textbf{UK Biobank~\cite{ukbiobank}:} The UK Biobank offers a vast dataset of genetic, phenotypic, and health-related information from 500,000 participants, supporting hypothesis generation in genomics and personalized medicine. Covering over 800 phenotypic traits and 96 million genetic variants, it enables causal inference and phenotypic prediction, evaluated through metrics like correlation coefficients, causal inference metrics, and prediction accuracy. While it excels in hypothesis validation, its focus is primarily on validating existing hypotheses rather than generating novel ones.

\textbf{MATBench~\cite{dunn2020benchmarking}:} MATBench is a benchmark dataset for material property prediction, featuring over 100,000 material samples across categories like alloys, ceramics, and polymers. It supports hypothesis generation about material properties and validates predictive models using metrics like prediction accuracy, RMSE, and \(R^2\). Its well-defined training, validation, and testing splits make it a robust resource for exploring novel material properties and systematically validating machine learning models in materials science.

\textbf{ClimateNet~\cite{prabhat2020climatenet}:} ClimateNet is an expert-labeled dataset designed for climate change research and hypothesis validation. It facilitates the identification and analysis of extreme weather patterns, enabling hypothesis generation in climate science by providing structured observational data. The dataset supports hypothesis validation by offering labeled climate events that can be used to train and test predictive models for extreme weather forecasting and climate anomaly detection. Key evaluation metrics include temporal trends analysis to assess long-term climate variations, anomaly detection for identifying deviations from expected climate behaviors, and spatial correlation metrics to measure the consistency of climate phenomena across regions. While ClimateNet is primarily used for validation rather than novel hypothesis discovery, its expert-curated labels and structured data representation make it a valuable tool for enhancing climate prediction models and improving the understanding of atmospheric processes.

\textbf{COCO Dataset~\cite{lin2014microsoft}:} The COCO (Common Objects in Context) dataset is a large-scale dataset designed for image captioning, object detection, and vision-language integration tasks. It enables hypothesis generation in artificial intelligence by providing a diverse set of annotated images for evaluating computer vision and natural language processing models. For example, a researcher might hypothesize that "transformer-based vision-language models generate more contextually accurate image captions than RNN-based models when multiple objects co-occur in complex scenes." This hypothesis can be tested using COCO's richly annotated images and corresponding captions. Hypothesis validation is supported through well-defined image-to-text relationships, allowing researchers to assess model performance in object recognition, scene understanding, and multimodal learning. Key evaluation metrics include precision and recall for object detection accuracy, BLEU score for measuring the alignment between generated and reference captions, and segmentation accuracy for validating instance-level object identification. While COCO primarily serves as a benchmarking dataset, its rich annotations and large-scale diversity facilitate advancements in AI-driven image interpretation and vision-language research.

\textbf{Gene Ontology (GO)~\cite{ashburner2000gene}:} The Gene Ontology (GO) dataset offers a hierarchical vocabulary for annotating gene products, supporting hypothesis generation and validation in genomics and molecular biology. It enables hypothesis generation by detailing gene functions, processes, and cellular components, and validates computational models for gene function prediction and pathway analysis. Key metrics include semantic similarity to measure functional similarities, graph-based validation using metrics like connectivity and path length, and annotation consistency to evaluate reliability. While GO is primarily used for hypothesis validation rather than novel discovery, its detailed annotations and hierarchical structure make it a robust tool for validating existing hypotheses.

\textbf{AHTech Electrolyte Additive Dataset~\cite{doi:10.1126/sciadv.adu4391}:} This dataset was introduced as part of the AHTech platform for accelerating electrochemical discovery. It contains high-throughput screening data from 180 small-molecule electrolyte additives tested for aqueous zinc metal batteries. Each additive was characterized across 200 electrochemical cycles to determine Coulombic efficiency, enabling the training of machine learning models for additive performance prediction. The dataset supports hypothesis validation by uncovering structure-performance relationships using techniques such as Shapley Additive Explanations (SHAP) and Spearman correlation. Its high experimental fidelity and structured annotations make it valuable for data-driven hypothesis generation in electrochemistry and battery materials research.

\textbf{CSKG-600~\cite{BORREGO2025113280}:} CSKG-600 is a benchmark dataset introduced to evaluate hypothesis generation over scholarly knowledge graphs. It consists of 600 candidate hypotheses manually labeled by domain experts as valid or invalid, supporting developing and evaluating link prediction models for scientific discovery. The dataset integrates structured triples with semantic and bibliometric metadata, enabling robust benchmarking of systems like ResearchLink. It facilitates validation through ranking-based metrics such as Precision@K and domain-specific agreement scores. As one of the first domain-independent resources in this space, CSKG-600 is well-suited for hypothesis validation tasks involving multi-modal, interdisciplinary scientific knowledge.

\textbf{DrugBank~\cite{wishart2018drugbank}:} DrugBank integrates comprehensive data on drugs and molecular targets, facilitating hypothesis generation and validation in pharmacology and bioinformatics. With over 14,000 drugs and 6,000 protein targets, it supports predictions in drug efficacy and adverse effects, evaluated through metrics like interaction accuracy, RMSE for binding affinity predictions, and drug-likeness scores. DrugBank excels in discovering novel drug candidates and provides experimental and clinical data for robust validation.

\textbf{AI2 Science Questions~\cite{clark2018think}:} The AI2 Science Questions dataset comprises over 10,000 multiple-choice questions, testing AI reasoning and knowledge representation in educational research. Covering topics from physics to earth science, it supports hypothesis generation about reasoning capabilities and evaluates model performance using metrics like accuracy, explainability scores, and logical consistency metrics. While it supports exploration of novel reasoning strategies in AI, it primarily focuses on evaluating reasoning rather than hypothesis feasibility.

\textbf{Materials Project~\cite{jain2013commentary}:} The Materials Project is a comprehensive dataset containing material properties and structures, facilitating hypothesis generation and validation in materials science. It supports the creation of hypotheses related to material properties and applications, validates predictive models in material discovery and design, and enables structure-property relationship analysis. Evaluation metrics include structural matching to validate predicted structures, novelty filtering to identify unique materials, and prediction accuracy to assess model reliability. This dataset fosters the discovery of novel materials through computational insights and provides experimental data to validate hypotheses, ensuring practical relevance.

\textbf{KEGG Pathway~\cite{kanehisa2000kegg}:} The KEGG Pathway dataset offers detailed insights into metabolic and signaling pathways, supporting hypothesis generation and validation in genomics and biomedicine. It facilitates the creation of hypotheses regarding biochemical interactions, validates predictive models for gene and protein interactions, and serves as a foundation for pathway enrichment analysis. With 540 curated pathways across metabolic, regulatory, and signaling processes, the dataset is regularly updated to reflect experimental and computational advances. Evaluation metrics include pathway enrichment scores, graph metrics (centrality, connectivity, modularity), and prediction accuracy. KEGG Pathway supports the discovery of novel biochemical relationships while providing a robust basis for hypothesis validation.

\textbf{American Community Survey (ACS)~\cite{acs}:} The American Community Survey (ACS) is an annual survey by the U.S. Census Bureau, providing comprehensive demographic, social, economic, and housing data. It facilitates hypothesis generation about trends, behaviors, and disparities, validates models for policy analysis and urban development, and offers longitudinal insights into population changes. Covering approximately 2.5 million households annually and featuring over 35,000 variables, ACS enables nationwide and local-level analysis. Metrics include statistical analysis, correlation coefficients for variable relationships, and subgroup comparisons. While not suited for novel hypothesis creation, the dataset’s representative and detailed nature ensures robust validation.

\textbf{Patent Data (USPTO)~\cite{uspto}:} The USPTO dataset contains extensive information on granted patents, offering a vital resource for hypothesis generation and validation in technology and innovation research. It facilitates the exploration of technological trends, patent networks, and collaboration patterns while validating hypotheses regarding novelty and impact. With over 10 million patents spanning industries, the dataset provides metadata like inventor details, filing dates, and citations. Key metrics include citation analysis (impact scores), novelty scores, and collaboration indices. The dataset excels in novelty analysis and trend identification but is less suited for direct feasibility testing.

\textbf{XSum~\cite{narayan2018don}:} The XSum dataset is a large-scale resource for abstractive text summarization, extensively used in NLP research. It supports hypothesis creation about summarization techniques, validates text generation frameworks, and benchmarks summarization algorithms. For example, a researcher might hypothesize that "pre-trained language models fine-tuned with reinforcement learning from human feedback (RLHF) produce more factually consistent summaries on XSum than models trained with maximum likelihood alone." This hypothesis can be evaluated using XSum's single-sentence human-written summaries and news articles across diverse topics. Comprising 226,711 news articles with single-sentence summaries across diverse topics, XSum offers rich contextual diversity. Metrics include BLEU for n-gram overlap, ROUGE for unigram and sequence comparisons, and conciseness metrics. The dataset fosters novel summarization methods and provides a robust foundation for validating advanced NLP models.

\textbf{Cosmic (Catalogue of Somatic Mutations in Cancer)~\cite{bamford2004cosmic}:} Cosmic is a comprehensive dataset detailing somatic mutations in cancer, supporting hypothesis generation and validation in cancer genomics and personalized medicine. It facilitates the exploration of genetic mutations linked to cancer progression, validates mutation prediction models, and serves as a foundation for studying mutation impacts across various cancer types. The dataset includes data on over 30,000 genes and 2 million somatic mutations, annotated with pathways, phenotypes, and clinical outcomes. Evaluation metrics such as mutation analysis metrics, pathway mapping metrics, and prediction accuracy validate hypotheses. While Cosmic excels in supporting hypothesis validation, it primarily focuses on existing hypotheses and provides robust, experimentally validated data for analysis.

\textbf{Open Research Knowledge Graph (ORKG)~\cite{jaradeh2019open}:} ORKG provides a structured, machine-readable representation of interdisciplinary research contributions, enabling hypothesis generation and validation. It uncovers relationships among research topics, datasets, and methods, validates hypotheses through graph-based analyses of citation impact and knowledge diffusion, and structures scientific knowledge for collaborative exploration. The dataset spans multiple disciplines, including AI, biology, and social sciences, with over 3 million triples linking research entities. Metrics such as graph centrality, novelty detection, and citation impact highlight its value in supporting novel connections and systematic validation.

\textbf{Conclusion.} The datasets presented in this section underscore their foundational role in hypothesis creation and validation across a wide range of scientific disciplines. Domain-specific datasets, such as \textit{PubMed Abstracts} and \textit{KEGG Pathway}, facilitate targeted research by providing structured knowledge for biomedical and genomic studies. \textit{Interdisciplinary datasets}, like the \textit{Open Graph Benchmark (OGB)} and \textit{Open Research Knowledge Graph (ORKG)}, enable hypothesis generation and validation across multiple domains, fostering cross-disciplinary innovation.  
Datasets emphasizing novelty and feasibility, such as \textit{Materials Project} and \textit{ChEMBL}, support scientific breakthroughs in materials science and drug discovery by offering rich, structured data for predictive modeling and validation. Resources like \textit{ClimateNet} and \textit{COCO Dataset} demonstrate how curated datasets drive advancements in climate science and AI-driven hypothesis evaluation.  
As these datasets continue to expand in scale, annotation quality, and accessibility, they will play an increasingly vital role in enhancing hypothesis generation, ensuring rigorous validation, and accelerating scientific discovery. By leveraging these diverse and evolving datasets, researchers can refine predictive models, validate novel hypotheses, and drive breakthroughs in emerging fields.
\section{Categorization of Hypothesis Generation Approaches}
\label{sec:hypothesis_generation}

Scientific hypothesis generation has been modeled through two primary paradigms: (1) computational frameworks for discovery grounded in symbolic reasoning and cognitive science, and (2) contemporary methods driven by large-scale neural models, particularly Large Language Models (LLMs). Each reflects a different intuition about how hypotheses are conceived, represented, and evaluated.
Computational frameworks for discovery view scientific hypothesis formation as a structured problem-solving process. Influenced by early work in cognitive science, these systems simulate how humans incrementally build explanations from observations by applying heuristic rules, constructing symbolic representations, and iteratively refining their models~\cite{bradshaw1983studying, simon1992scientific, langley1988computational, langley1998computer, langley2000computational, dvzeroski2007computational, langley2013central, langley2024integrated}. Tools such as \textit{BACON}~\cite{bradshaw1983studying} and \textit{KEKADA}~\cite{langley2000computational} exemplify this approach by rediscovering known laws and relationships in structured datasets. These methods define a hypothesis space $\mathcal{H}$ as a set of symbolic expressions—such as Newton's second law or Mendelian inheritance rules—generated through grammars, algebraic forms, or logic-based templates, often constrained by background knowledge or domain-specific primitives. Candidate hypotheses within this space are constructed using heuristic search strategies such as forward chaining~\cite{langley2024integrated}, rule induction, or equation synthesis. They are ranked based on their empirical fit to observed data and structural simplicity, typically favoring parsimonious and generalizable formulations. This process is often formalized as a weighted scoring function:
\begin{equation}
\text{score}(h) = \alpha \cdot \text{fit}(h, D) - \beta \cdot \text{complexity}(h),
\end{equation}
where $h \in \mathcal{H}$ is a hypothesis, $D$ is the dataset, and $\alpha$, $\beta$ are weights balancing empirical accuracy and parsimony. 
To illustrate, consider the task of rediscovering the Hall-Petch relationship in materials science, which relates the yield strength $\sigma_y$ of a polycrystalline material to its grain size $d$ through the equation $\sigma_y = \sigma_0 + k \cdot d^{-1/2}$. A candidate hypothesis in this domain may take the form:
$h(d) = a + b \cdot d^{-c},$
where $a$, $b$, and $c$ are free parameters to be estimated from data. The empirical fit can be computed using mean squared error:
$\text{fit}(h, D) = - \frac{1}{N} \sum_{i=1}^N \left( h(d_i) - \sigma_{y_i} \right)^2,$
where $(d_i, \sigma_{y_i}) \in D$ are observed grain size and yield strength pairs. The complexity of $h$ can be quantified by counting the number of mathematical operators and the depth of the expression tree:
$\text{complexity}(h) = \text{NumOperators}(h) + \lambda \cdot \text{TreeDepth}(h).$
This formalism favors hypotheses that not only fit the data well but are also structurally simple and generalizable, reflecting core principles of scientific discovery.
These approaches are particularly valuable when working with well-structured, interpretable data and when the hypothesis space is tightly constrained by theory. However, their reliance on handcrafted rules and domain-specific encodings limits their scalability and effectiveness in data-rich, unstructured, or ambiguous environments.

While symbolic frameworks provide a principled and interpretable foundation for modeling hypothesis generation, they often rely on domain-specific encodings, constrained rule spaces, and hand-crafted heuristics. As scientific data's scale, heterogeneity, and ambiguity have increased, these limitations have spurred the adoption of more flexible and data-driven approaches. This shift has been catalyzed by the emergence of Large Language Models (LLMs), which enable hypothesis generation by synthesizing knowledge across unstructured sources at scale. Unlike symbolic methods that construct hypotheses from explicitly defined rule spaces, LLMs operate over implicit probabilistic representations learned from diverse corpora, offering new possibilities for discovery in underexplored or interdisciplinary domains. These models are trained on massive, heterogeneous corpora that span scientific literature, code repositories, and multimodal sources. Rather than relying on symbolic reasoning, LLMs use statistical learning to approximate the conditional probability of generating a hypothesis $H$ given a context $C$, typically modeled as:
\begin{equation}
P(h \mid c) = \prod_{t=1}^{T} P(h_t \mid h_{<t}, c; \theta),
\end{equation}
where $h = (h_1, \ldots, h_T)$ is a sequence of tokens representing a candidate hypothesis, and $\theta$ are the model parameters learned from data. LLMs excel at generalizing across domains, synthesizing knowledge from unstructured input, and proposing hypotheses that may span disciplinary boundaries. They are particularly effective in cases where structured models are not readily available or where rapid exploration of diverse ideas is desired. However, LLMs often lack formal mechanisms for explanation, causality, and logical rigor, necessitating downstream validation via simulation, symbolic reasoning, or expert review~\cite{wang2024dspybasedneuralsymbolicpipelineenhance}.

\begin{table}[ht]
\centering
\caption{Comparison of Symbolic and LLM-Based Hypothesis Generation Approaches}
\begin{tabular}{p{2.6cm}|p{5cm}|p{5cm}}
\hline
\textbf{Aspect} & \textbf{Symbolic Discovery Systems} & \textbf{LLM-Based Generative Systems} \\ \hline

Example Systems & BACON~\cite{bradshaw1983studying}, KEKADA~\cite{langley2000computational} & ChatGPT~\cite{achiam2023gpt}, SciAgents~\cite{ghafarollahi2024sciagents} \\ \hline

Hypothesis Space Construction & Explicit, rule-based; defined using symbolic grammars, logic rules, and algebraic templates & Implicit, data-driven; encoded via pretraining on large corpora and fine-tuning for specific domains \\ \hline

Inference Mechanism & Heuristic or search-based (e.g., forward/backward chaining, ILP) & Generative decoding using probabilistic token prediction and attention-based reasoning \\ \hline

Validation Strategy & Fit to empirical data, parsimony (Occam’s razor), consistency with domain theory & Rationale scoring, entailment verification, retrieval-augmented consistency checking \\ \hline

Interpretability & High (transparent rule structures, derivations can be traced) & Low to moderate (requires explainability tools such as SHAP~\cite{lundberg2017shap}, LIME~\cite{ribeiro2016lime}, or prompt engineering) \\ \hline

Scalability & Limited by combinatorial rule space and symbolic inference complexity & High scalability due to pretraining and fine-tuning across large-scale unstructured datasets \\ \hline

Typical Application Domains & Classical scientific discovery (e.g., physics, chemistry, cognitive modeling) & Interdisciplinary science, biomedical literature mining, materials science, automated experimentation \\ \hline

Strengths & Theory-grounded, interpretable, robust in structured domains & Flexible, cross-domain generalization, effective in handling unstructured or sparse data \\ \hline

Limitations & Labor-intensive to construct, brittle with noisy data, domain-dependent & Prone to hallucination, limited interpretability, sensitive to prompt and data biases \\ \hline

\end{tabular}
\label{tab:symbolic_vs_llm}
\end{table}

The choice between symbolic and LLM-based approaches depends largely on the nature of the task and the structure of available data. Symbolic frameworks are most effective when the objective is to derive interpretable models from well-defined variables or to formulate hypotheses aligned with established scientific theories. In contrast, LLMs are well-suited for contexts involving large-scale, unstructured datasets or where the discovery of novel, cross-domain associations is prioritized. Increasingly, hybrid pipelines that combine LLM-driven generation with symbolic or simulation-based validation are being adopted to balance generative flexibility with interpretability and rigor~\cite{langley2024integrated, ghafarollahi2024protagents, ren2025scientificintelligencesurveyllmbased}. Recent advancements exemplify this convergence: the AHTech platform~\cite{doi:10.1126/sciadv.adu4391} integrates automated electrochemical experimentation with machine learning to enable high-throughput hypothesis testing in battery research; LLMs have been shown to automate bioinformatics workflows when paired with structured repositories like cBioPortal~\cite{10765518}; and modular systems modeling frameworks such as Robotics-LLM~\cite{Yin_Feng_Bao_Gao_Liang_Heather_Aspuru-Guzik_Wang_2025} and proteomics-based KDD pipelines~\cite{Resell2025.02.23.639474} facilitate hypothesis refinement in chemical discovery and oncology, respectively. These efforts underscore modern hypothesis generation evolving into a multifaceted, interdisciplinary process where automation, experimentation, and semantic reasoning coalesce to accelerate discovery. They also highlight the growing need for domain-specific, interpretable, and safety-aware AI agents~\cite{ren2025scientificintelligencesurveyllmbased, yu2025ai, steinecker2025towards}.

Building on these foundations, the remainder of this section categorizes contemporary hypothesis generation approaches into distinct methodologies. These include knowledge-driven methods, data-driven integration, AI-driven exploration, text and concept mining, simulation and modeling, interactive and collaborative systems, causal inference, dynamic and adaptive knowledge systems, and multi-agent systems. Each approach reflects a unique computational intuition and contributes to the evolving landscape of scientific discovery through its specialized techniques and domain applications.
By showcasing these methodologies, we aim to highlight their transformative potential in fostering novel and impactful scientific discoveries. Figure~\ref{fig:hypothesis_generation_pipeline} presents a hypothesis generation pipeline highlighting various methods that contribute to producing candidate hypotheses that are novel and plausible.

\begin{figure*}[ht] 
    \centering
    \includegraphics[width=\textwidth]{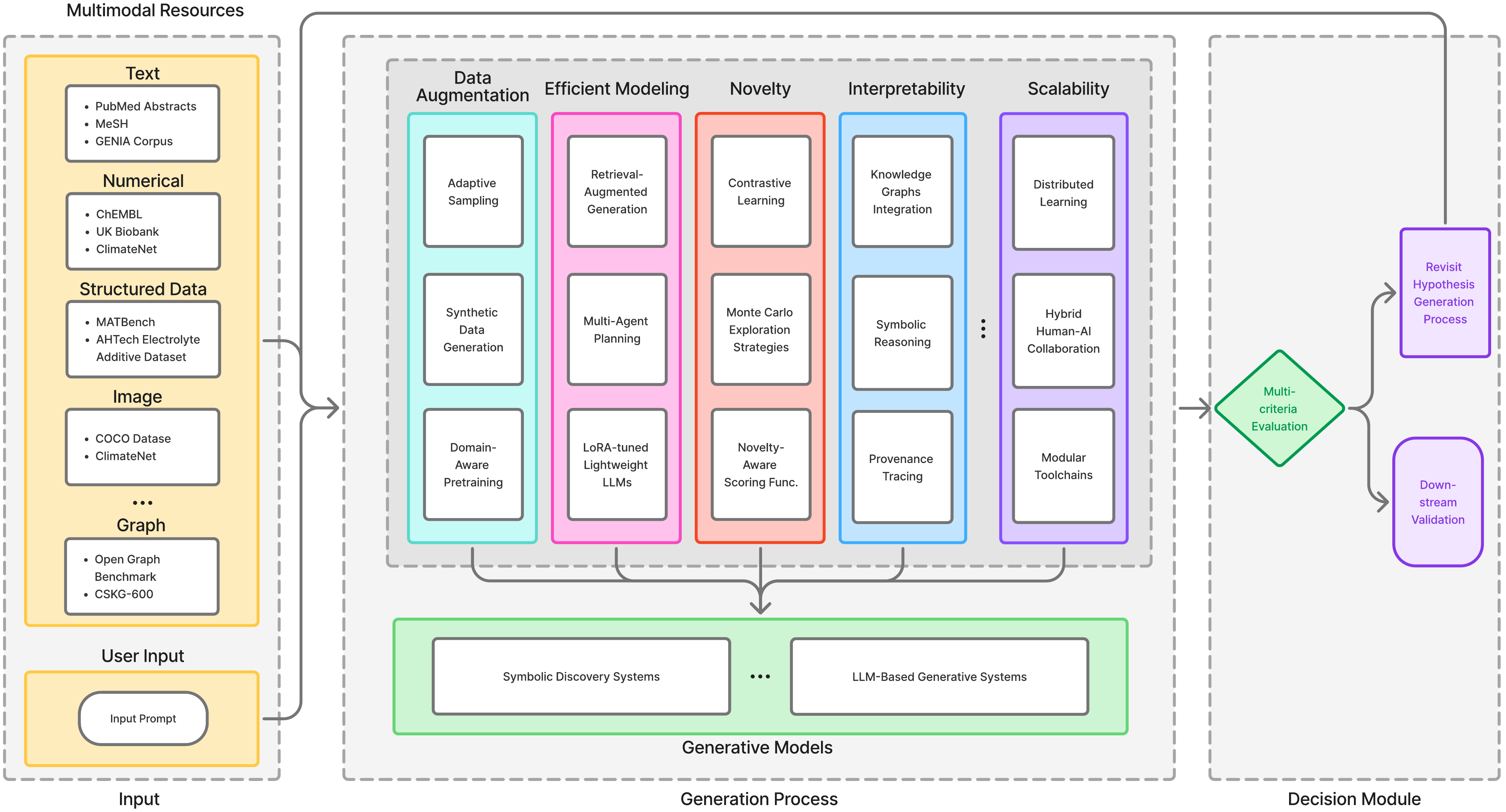}
    \caption{Modular pipeline for AI-driven hypothesis generation. The figure illustrates how multimodal data sources flow through symbolic and LLM-based generative components, incorporating retrieval, reasoning, scoring, and refinement to support interpretable and novelty-aware hypothesis generation.}   
    \label{fig:hypothesis_generation_pipeline}
\end{figure*} 

\begin{table}[ht]
\centering
\caption{Summary of Hypothesis Generation Approaches and Tools}
\begin{tabular}{p{2cm}|p{2cm}|p{2cm}|p{2.7cm}|p{2.7cm}}
\hline
\textbf{Approach} & \textbf{Example Tool/System} & \textbf{Domain} & \textbf{Strengths} & \textbf{Weaknesses} \\ \hline
Knowledge-Driven Methods & MOLIERE~\cite{sybrandt2018large} & Biomedicine, Interdisciplinary & Ensures consistency and logical validity & Limited novelty, relies on existing knowledge \\ \hline
Data-Driven Integration & SciAgents~\cite{ghafarollahi2024sciagents} & Biomedicine, Scientific Discovery & Merges structured and unstructured data & Quality depends on dataset reliability \\ \hline
AI-Driven Exploration & Reinforcement Learning~\cite{zhou2024hypothesisgenerationlargelanguage} & Drug Discovery, Materials Science & Uncovers novel patterns beyond human intuition & High computational cost, lacks interpretability \\ \hline
Text and Concept Mining & Chemist-X~\cite{chen2024chemist} & Chemistry, Biomedical Sciences & Extracts insights from large-scale text data & Sensitive to NLP noise, domain adaptation required \\ \hline
Simulation and Modeling & VELMA~\cite{schumann2024velma} & Robotics, Engineering & Models diverse scenarios and unconventional ideas & Requires high computational resources \\ \hline
Interactive and Collaborative Systems & Human-AI Collaboration~\cite{kim2018research} & Education, Ethics & Combines human insights with AI efficiency & Dependent on human oversight, potential biases \\ \hline
Causal Inference & Causal Discovery~\cite{jha2019hypothesis} & Biomedicine, Social Science & Identifies causal relationships, enhances reliability & Requires large datasets, vulnerable to confounders \\ \hline
Dynamic and Adaptive Knowledge Systems & Knowledge Graph Updates~\cite{beltagy2019scibert} & Pharmacology, AI Research & Continuously refines and contextualizes knowledge & High processing power needed for real-time updates \\ \hline
Multi-Agent Systems & Multi-Agent AI~\cite{qi2024large} & Genomics, Collaborative Science & Enables decentralized, expert-driven collaboration & Complexity increases with agent interactions \\ \hline
\end{tabular}
\label{tab:hypothesis_summary}
\end{table}

\subsection{Knowledge-Driven Methods}

Knowledge-driven methods, which include knowledge graphs, network-based approaches, and ontology-based reasoning, represent a cornerstone of modern hypothesis generation. These methods provide a systematic and structured way to explore scientific knowledge's vast and often overwhelming complexity. Organizing information into structured representations enables researchers to uncover hidden patterns, establish interdisciplinary connections, and generate innovative and contextually relevant hypotheses. These approaches are particularly valuable in domains where the complexity of interactions—such as between genes, proteins, diseases, or materials—defies traditional manual analysis.

Knowledge graphs serve as powerful tools for capturing and visualizing relationships between entities. Knowledge graphs facilitate intuitive exploration and semantic reasoning by structuring knowledge as nodes (representing entities such as genes, diseases, or chemical compounds) and edges (representing their relationships). Systems like MOLIERE leverage vast biomedical repositories such as PubMed to identify novel gene-disease associations that often elude conventional analysis~\cite{sybrandt2018large}. Similarly, SciAgents demonstrates the integration of dynamic knowledge graphs with large language models (LLMs), enabling interdisciplinary research in fields like pharmacology, where complex relationships must be navigated to propose innovative hypotheses~\cite{ghafarollahi2024sciagents}. These systems allow researchers to uncover overlooked connections, prioritize promising research directions, and bridge gaps across disciplines, accelerating discovery. Recent work has also explored the reverse paradigm: using LLMs themselves as latent knowledge graphs. Instead of explicitly constructing graph structures, these approaches treat LLMs as implicit knowledge stores capable of retrieving, composing, and reasoning over relational information. For instance, \cite{kau2024combining} proposes prompting strategies and architectural mechanisms that enable LLMs to emulate knowledge graph behavior by performing multi-hop reasoning or generating structured triples directly from natural language inputs. While this direction offers scalability and ease of deployment advantages, reducing the need for manual graph construction, it often sacrifices interpretability, semantic precision, and logical consistency. LLM-based representations lack the explicit structure and auditability of formal knowledge graphs, making them less suitable for domains requiring rigorous semantic alignment. Therefore, these approaches are best viewed as complementary; LLMs can enhance knowledge graph population, reasoning, and hypothesis suggestion, but structured graphs remain crucial for precise and semantically grounded scientific hypothesis generation.

A key strength of knowledge graphs lies in their ability to perform advanced analyses such as link prediction and graph-based learning. Link prediction, which leverages metrics like embeddings, centrality, and similarity scores, has proven particularly impactful in areas like drug discovery, where identifying new drug-disease interactions can expedite therapeutic development~\cite{kim2018research}. Recent advancements, such as Graph Neural Networks (GNNs), further extend the power of knowledge graphs by modeling intricate, high-dimensional relationships~\cite{bai2024dynamic}. GNNs excel in revealing latent patterns and dependencies, enabling the analysis of multi-layered interactions that drive innovation in fields like materials science, environmental modeling, and beyond~\cite{beltagy2019scibert}. Furthermore, the dynamic nature of modern knowledge graphs ensures their adaptability, as they can incorporate newly available data to refine their structure and maintain relevance in rapidly evolving scientific landscapes~\cite{qi2024large}.

Ontology-based reasoning complements knowledge graphs by introducing formalized frameworks for semantic consistency and logical reasoning. Unlike knowledge graphs, which focus on uncovering relationships, ontology-based systems emphasize ensuring that generated hypotheses adhere to established domain-specific standards and terminologies. These systems use well-defined ontologies to capture classes, properties, and relationships, thereby enabling precise and semantically valid hypothesis generation. For example, ontology-based hypothesis validation has been used to identify novel gene-disease relationships by mapping genetic functions to clinical phenotypes. The integration of multiple ontologies has further unified knowledge across domains such as physics, biology, and chemistry, facilitating interdisciplinary research~\cite{wang2023scimon}.

The methodologies underpinning ontology-based reasoning include ontology mapping, semantic inference, and automated reasoning. Ontology mapping aligns diverse ontologies from multiple domains, enabling unified views of knowledge essential for interdisciplinary research~\cite{kim2018research}. Semantic inference, which applies logical rules to infer new relationships, uncovers insights that are otherwise difficult to detect. Automated reasoning tools derive novel hypotheses while maintaining consistency with existing knowledge structures~\cite{beltagy2019scibert}. These techniques have demonstrated significant utility in biomedical science~\cite{10765518}, drug discovery~\cite{blanco2023role}, and interdisciplinary innovation~\cite{zhang2024decoupled}, with applications ranging from identifying novel drug targets to exploring complex ecological systems~\cite{qi2024large}.

Together, knowledge graphs and ontology-based reasoning form a complementary toolkit for hypothesis generation. Knowledge graphs~\cite{bai2024advancing} excel at uncovering patterns and relationships that are often hidden in vast, unstructured datasets, while ontology-based reasoning~\cite{liu2010ontology} ensures that hypotheses are both precise and semantically consistent. This synergy has profound implications across disciplines. In biomedicine, these methods have mapped intricate relationships between genes, diseases, and drugs, facilitating breakthroughs in personalized medicine and rare disease research~\cite{ kim2018research}. In materials science, they have accelerated the discovery of novel materials by identifying promising chemical combinations~\cite{ghafarollahi2024protagents}. Environmental science has also benefited from these approaches, with knowledge graphs uncovering correlations between oceanic and atmospheric variables to model ecological phenomena~\cite{wang2023scimon}. Ontology-based reasoning has further strengthened interdisciplinary research by aligning diverse terminologies and frameworks, enabling seamless collaboration and hypothesis generation across fields.

While knowledge-driven methods have proven transformative, they are not without limitations. Knowledge graphs rely heavily on structured data, which can constrain their ability to generate truly novel hypotheses when data is incomplete or biased. Similarly, ontology-based reasoning is limited by the granularity and completeness of the underlying ontologies. To address these challenges, researchers are increasingly integrating dynamic updates and machine learning techniques. For instance, GNNs and other machine learning models enhance the ability of knowledge graphs to explore higher-order interactions and adapt to new data. Ontology-based systems are also evolving, incorporating learning-based methods to identify less obvious relationships and expand their scope. These advancements ensure that knowledge-driven methods remain adaptable, innovative, and capable of addressing the complexities of modern scientific inquiry.

In conclusion, knowledge-driven methods are indispensable for navigating and extracting value from the vast complexity of scientific knowledge. By combining structured representations, semantic rigor, and dynamic adaptability, they provide researchers with the tools to uncover hidden patterns, bridge disciplines, and generate actionable hypotheses. As these methods continue to evolve, their potential to drive innovation and foster interdisciplinary collaboration remains boundless.

\subsection{Data-Driven Integration}

Data-driven integration represents a transformative approach to hypothesis generation by synthesizing diverse datasets from disciplines such as genomics, proteomics, environmental science, and beyond. This methodology bridges traditionally isolated domains, uncovering complex relationships and enabling researchers to address multifaceted scientific challenges with unprecedented precision. By leveraging cross-domain data integration, researchers can construct a holistic understanding of intricate systems, fostering innovative solutions and advancing scientific discovery across a wide array of fields.

At the heart of data-driven integration lie methodologies designed to analyze and synthesize heterogeneous datasets, revealing insights that are otherwise inaccessible. Multi-omics integration, for instance, combines data from genomics, transcriptomics, and proteomics to identify high-value targets such as gene-protein interactions~\cite{qi2024large}. This approach has driven breakthroughs in personalized medicine by uncovering molecular mechanisms underlying diseases and identifying actionable therapeutic targets. Beyond omics, interdisciplinary integration extends these principles to link disparate datasets, such as combining genomic information with environmental data to explore broader patterns and emergent properties~\cite{pammi2023multiomics}. These techniques often rely on systems-level computational models, which simulate interactions across disciplines, providing a dynamic framework for hypothesis generation~\cite{kim2018research}. By capturing non-obvious correlations and emergent behaviors, these models empower researchers to explore complex phenomena that transcend traditional disciplinary boundaries.

A growing suite of tools exemplifies the far-reaching impact of data-driven integration in scientific hypothesis generation across diverse domains. In agriculture, VirtualPlant~\cite{katari2010virtualplant} integrates genomic, transcriptomic, and phenotypic data to uncover genetic pathways that enhance crop resistance to environmental stressors, supporting sustainable farming practices~\cite{qi2024large, kim2018research}. In pharmacology, BioLunar~\cite{wysocki2024llm} leverages multi-omics data to elucidate complex gene-protein interactions, enabling the discovery of precision therapies and accelerating the identification of drug-protein interactions~\cite{kim2018research, qi2024large}. In environmental science, tools like Climate KG~\cite{wu2022linkclimate} link climate variables with ecological and biological datasets to uncover factors driving ecosystem resilience and inform conservation and adaptation strategies. These examples not only demonstrate the versatility of integrative tools in enabling hypothesis generation but also highlight their broader utility in addressing critical challenges across genomics, pharmacology, agriculture, and environmental science.

The novelty of data-driven integration lies in its ability to connect datasets traditionally analyzed in isolation. For example, integrating ocean salinity data with atmospheric pressure measurements has provided new insights into climate dynamics that were previously unattainable~\cite{wang2023scimon}. However, this potential for discovery is contingent on the availability of high-quality datasets and the computational resources required to process and interpret them effectively. Resource constraints and dataset heterogeneity pose challenges that must be addressed to fully realize the promise of data-driven integration~\cite{gruver2024fine}. To overcome these barriers, adaptive algorithms, reinforcement learning, and dynamic systems modeling are increasingly employed. These methods not only enhance computational efficiency but also enable the exploration of deeper, non-obvious correlations within interdisciplinary datasets~\cite{zhou2024hypothesisgenerationlargelanguage}.

Data-driven integration serves as a powerful paradigm for addressing complex scientific questions, bridging disciplinary divides, and fostering novel insights. As computational methods and data quality continue to improve, this approach is poised to redefine the landscape of hypothesis generation. By synthesizing diverse data sources into cohesive frameworks, data-driven integration empowers researchers to tackle intricate challenges and unlock discoveries that were once beyond reach. The continued evolution of these methodologies promises to drive progress across a wide spectrum of fields, solidifying their role as a cornerstone of modern scientific inquiry.

\subsection{AI-Driven Exploration}

AI-driven exploration represents a transformative paradigm in hypothesis generation, integrating advanced methodologies such as machine learning (ML), statistical modeling, retrieval-augmented generation (RAG), and reinforcement learning (RL). These approaches empower researchers to process vast and complex datasets, synthesize real-time information, and navigate intricate hypothesis spaces to uncover innovative solutions. By leveraging adaptive algorithms, dynamic retrieval mechanisms, and iterative refinement strategies, AI-driven methods address the challenges of modern scientific inquiry across fields such as drug discovery, materials science, robotics, and social sciences~\cite{qi2024large, gruver2024fine}.

RAG exemplifies the dynamic capabilities of AI in hypothesis generation by combining the language comprehension power of LLMs with domain-specific datasets and real-time data retrieval systems. This approach bridges the gap between static repositories and evolving research challenges, ensuring that hypotheses are timely and grounded in relevant data~\cite{zhou2024hypothesisgenerationlargelanguage}. Key techniques in RAG include dynamic information retrieval, prompt engineering, and adaptive memory integration. Dynamic retrieval enables the extraction of up-to-date information from diverse sources, enriching hypotheses with the latest knowledge~\cite{beltagy2019scibert}. Carefully crafted prompts guide LLMs to generate contextually specific outputs, addressing discipline-specific challenges with precision~\cite{kim2018research}. Memory integration further enhances RAG’s capabilities by incorporating feedback from previous outputs, fostering coherence and novelty~\cite{qi2024large}. Tools like Chemist-X and SciAgents exemplify RAG’s versatility, applying it to drug discovery, interdisciplinary research, and pharmacology~\cite{chen2024chemist, ghafarollahi2024sciagents}.

RL adds another dimension to AI-driven exploration by leveraging trial-and-error strategies to refine hypotheses dynamically. RL systems optimize solutions in high-dimensional and non-linear hypothesis spaces, enabling the discovery of novel insights and the refinement of processes~\cite{fawzi2022discovering}. Techniques such as policy optimization iteratively improve decision-making processes, while model-free RL approaches like Q-learning facilitate hypothesis generation in environments with complex or poorly understood dynamics~\cite{gruver2024fine}. Multi-agent RL systems expand this capability further by employing collaborative agents to explore diverse hypothesis spaces simultaneously, fostering interdisciplinary insights and significantly increasing efficiency~\cite{zhou2024hypothesisgenerationlargelanguage}.

Several tools showcase the practical applications of these methodologies. VELMA demonstrates the integration of RAG with navigation and robotics, using textual and visual data to hypothesize optimal strategies for urban navigation~\cite{schumann2024velma}. Chemist-X applies RAG to chemical databases, generating reaction pathways and identifying novel drug candidates~\cite{chen2024chemist}. In RL, tools like RL-Discovery optimize material properties by exploring hypothesis spaces, accelerating advancements in catalyst design~\cite{gruver2024fine}. DrugRL leverages Q-learning to refine molecular configurations in drug discovery, while SciAgents integrates RL and knowledge graph approaches to enhance hypothesis validation across interdisciplinary fields~\cite{ghafarollahi2024sciagents}.

AI-driven exploration has demonstrated transformative impact across scientific disciplines. In drug discovery, RL frameworks such as DrugRL have accelerated the design of molecular structures, reducing costs and timelines by iteratively refining candidate molecules~\cite{qi2024large}. In materials science, RL-Discovery has optimized catalysts and identified compounds with unique properties, addressing critical performance criteria~\cite{gruver2024fine}. RAG tools like SciAgents have facilitated interdisciplinary queries, enabling researchers to tackle biomedical and computational challenges through systematic hypothesis refinement~\cite{ghafarollahi2024sciagents}. These examples highlight the adaptability and efficacy of AI-driven exploration in addressing complex, multi-dimensional problems.

While these methodologies hold immense promise, they are not without challenges. RAG systems depend on the quality and diversity of retrieved data, and poorly curated sources can constrain the novelty of hypotheses~\cite{beltagy2019scibert}. RL systems, meanwhile, are heavily influenced by the design of reward functions, with poorly defined rewards potentially leading to suboptimal exploration~\cite{fawzi2022discovering}. To address these limitations, researchers are integrating complementary approaches such as Bayesian optimization and adaptive memory mechanisms to balance exploration and exploitation while enhancing creativity and scientific relevance~\cite{zhou2024hypothesisgenerationlargelanguage}.

In conclusion, AI-driven exploration integrates RAG, RL, and advanced ML techniques into a cohesive framework for hypothesis generation. By navigating dynamic data landscapes, refining hypotheses iteratively, and uncovering high-value discoveries, these methodologies are expanding the frontiers of scientific inquiry. As AI-driven systems continue to evolve, their potential to foster interdisciplinary collaboration, drive innovation, and accelerate breakthroughs across diverse fields remains boundless.

\subsection{Text and Concept Mining}

Text and concept mining have emerged as indispensable methodologies for extracting meaningful insights from vast repositories of unstructured textual data, such as scientific literature, patents, and technical reports. These approaches systematically analyze and track the evolution of concepts, enabling researchers to uncover latent patterns, identify emerging trends, and generate hypotheses that align with the ever-changing knowledge landscape. By bridging information across disciplines and highlighting connections that may otherwise remain obscured, text and concept mining empower scientific discovery and innovation~\cite{pajo2025leveraging}.

At the core of text and concept mining are advanced techniques designed to distill valuable insights from massive text corpora. Topic modeling algorithms, such as Latent Dirichlet Allocation (LDA), cluster related information into coherent themes, revealing underlying patterns that might not be immediately apparent~\cite{beltagy2019scibert}. Dynamic evolution models extend these capabilities by analyzing temporal changes in text representations, such as word embeddings or semantic networks, to capture shifts in research priorities and conceptual frameworks~\cite{qi2024large}. These longitudinal analyses are critical for identifying emerging fields of study and generating hypotheses that reflect current and future trends. Natural Language Processing (NLP) pipelines further enhance the process by automating tasks like text preprocessing, feature extraction, and semantic analysis, ensuring that the generated hypotheses are both contextually relevant and actionable~\cite{zhou2024hypothesisgenerationlargelanguage}.

A range of innovative tools exemplifies the transformative potential of text and concept mining in hypothesis generation. Dyport combines text mining with dynamic graph modeling to uncover temporal trends and co-occurrence networks in genomics, enabling the identification of novel gene-disease associations~\cite{tyagin2024dyport}. SciFact leverages co-occurrence patterns in scientific literature to propose new drug-disease relationships, opening avenues for drug repurposing. ConceptNet is a pre-built common-sense knowledge base, helping researchers identify knowledge gaps and propose innovative hypotheses in diverse fields~\cite{kim2018research}. These tools collectively demonstrate how text and concept mining methodologies can address complex scientific challenges by providing a dynamic framework for discovery.

The applications of text and concept mining span a wide array of scientific disciplines, illustrating their versatility and impact. In genomics, Dyport has analyzed temporal trends in gene-variant literature, revealing emerging links to rare diseases and prioritizing research directions~\cite{tyagin2024dyport}. In drug discovery, SciFact has enabled the rapid identification of novel drug-disease pairings through large-scale analysis of biomedical texts, significantly accelerating the exploration of repurposing opportunities. In climate science, dynamic evolution models have tracked decades of literature to highlight shifts in research priorities, methodologies, and key focus areas, offering critical insights into the evolution of the field. These examples underscore the adaptability of text and concept mining in addressing interdisciplinary and domain-specific challenges.

The novelty of hypotheses generated by text and concept mining stems from their ability to uncover emerging connections and align with cutting-edge research directions. By analyzing temporal trends and co-occurrence patterns, these methods reveal hidden insights and suggest promising avenues for exploration. However, their effectiveness depends on the quality, recency, and completeness of the underlying datasets. Outdated or incomplete text corpora can constrain the scope and relevance of generated hypotheses, limiting their potential impact. To overcome these limitations, advancements such as multilingual and multimodal text mining, which combine textual data with other modalities such as images, tables, or graphs to uncover richer insights, are being integrated, broadening the diversity of insights and fostering cross-disciplinary hypothesis generation. These enhancements ensure that text and concept mining remain robust tools for driving scientific innovation.

By harnessing the vast and ever-expanding repositories of textual data, text and concept mining provide researchers with a powerful, adaptive framework for hypothesis generation. Their ability to track the evolution of knowledge, uncover emerging trends, and propose novel connections positions them as essential methodologies in the modern scientific toolkit. As these techniques continue to evolve, their role in advancing research across disciplines is poised to expand, catalyzing progress and innovation in diverse scientific domains.

\subsection{Simulation and Modeling}

Simulation and modeling approaches~\cite{schumann2024velma, lavin2021simulation,sarathy2022biplex,clark2020simulating,helie2008knowledge} are revolutionizing hypothesis generation by mimicking creative problem-solving processes to explore unconventional ideas and solutions. These techniques leverage advanced computational frameworks to emulate divergent thinking, allowing researchers to transcend traditional methodologies and tackle complex challenges with fresh perspectives. By enabling exploration across vast solution spaces, simulation and modeling foster innovation and uncover opportunities that might otherwise remain hidden. This approach has proven particularly effective in addressing non-linear, multi-dimensional problems across diverse scientific disciplines~\cite{kim2018research, qi2024large}.

Central to simulation and modeling are techniques designed to simulate creativity and enhance exploratory reasoning. Simulated annealing, inspired by natural processes of energy optimization, iteratively balances exploration and exploitation to identify optimal hypotheses. This method ensures thorough exploration of solution spaces while converging toward high-value outcomes. Creative neural networks, leveraging specialized architectures, emulate human-like reasoning and imagination to propose novel and unconventional hypotheses. These networks excel in synthesizing innovative ideas, making them particularly valuable in domains where groundbreaking insights are essential. Idea mining, which combines text mining and clustering techniques, extracts unique concepts from large datasets, enhancing the diversity and originality of generated hypotheses. Together, these methodologies provide a powerful framework for generating novel and impactful ideas.

A variety of cutting-edge tools demonstrate the transformative potential and practical versatility of simulation and modeling in hypothesis generation across diverse domains. In the social sciences, SimHypoth~\cite{lin2023introduction} employs neural network-based models to uncover non-linear relationships and generate hypotheses that challenge conventional assumptions, including correlations between demographic factors and societal trends. In engineering and materials science, IdeaFlow~\cite{utley2022ideaflow} tracks the diversity of ideas emerging from simulations, enabling the discovery of innovative design strategies and material combinations that have advanced sustainable infrastructure, including novel approaches to bridge and building construction. In theoretical physics, GenCreative~\cite{de2017learning} leverages generative adversarial networks (GANs) to explore unconventional solutions to unsolved problems, from subatomic particle interactions to alternative theoretical frameworks, thereby expanding both theoretical inquiry and experimental frontiers. Collectively, these tools exemplify how simulation and modeling techniques can drive scientific innovation and address complex, interdisciplinary challenges.

The novelty of hypotheses generated through simulation and modeling is one of their defining strengths. By embracing unconventional approaches, these techniques frequently lead to insights that challenge existing paradigms and expand the boundaries of current knowledge. For instance, IdeaFlow proposed novel material combinations in construction that had not been previously considered, demonstrating the power of creative exploration~\cite{kim2018research}. However, the reliance on randomness or unstructured exploration in some simulations can occasionally produce impractical or untestable hypotheses, limiting their utility. To mitigate this, researchers increasingly incorporate domain-specific constraints and iterative feedback loops, ensuring that hypotheses remain both innovative and feasible. These enhancements balance creativity with relevance, allowing simulation and modeling to generate actionable insights.

Simulation and modeling represent a dynamic frontier in hypothesis generation, blending computational power with creativity to tackle complex scientific challenges. By fostering unconventional thinking and expanding solution spaces, these approaches provide a robust framework for advancing scientific inquiry and driving innovation across disciplines. As techniques and tools continue to evolve, their capacity to generate transformative insights will remain central to addressing the complexities of modern research.

\subsection{Interactive and Collaborative Systems}

Interactive and collaborative systems, including human-in-the-loop (HITL) frameworks, combine the intuitive reasoning of human experts with the computational power of AI, creating a synergistic approach to hypothesis generation. By integrating iterative feedback from users, these systems refine AI-generated outputs to ensure hypotheses are relevant, innovative, and aligned with real-world contexts. This blend of human creativity and machine efficiency enables researchers to explore complex problems, bridging gaps between intuition-driven insights and data-driven precision~\cite{kim2018research, ghafarollahi2024sciagents}.

The methodologies driving interactive and collaborative systems focus on iterative learning, crowdsourced contributions, and explainability. Iterative learning fosters a continuous feedback loop between human users and AI, allowing hypotheses to be refined through multiple cycles of interaction. This ensures that the outputs align with the strategic priorities and nuanced understanding of domain experts~\cite{zhou2024hypothesisgenerationlargelanguage}. Crowdsourced contributions expand this collaborative framework by aggregating input from diverse participants, enriching the hypothesis space with multiple perspectives and enhancing robustness~\cite{kim2018research}. Explainability mechanisms ensure that AI-generated hypotheses are interpretable, providing transparent justifications for their derivation. This fosters trust and enables meaningful feedback, ultimately improving the quality and credibility of the generated hypotheses~\cite{beltagy2019scibert}.

A variety of tools exemplify the practical implementation of these methodologies, showcasing their versatility across scientific domains. SciAgents integrates real-time expert feedback with AI-driven hypothesis generation, enabling the refinement of complex hypotheses in biomedical and pharmacological research~\cite{ghafarollahi2024sciagents}. CrowdScience leverages crowdsourcing to generate hypotheses on behavioral and social phenomena, aggregating insights from a wide participant pool to ensure inclusivity and diversity. ExplanatoryAI focuses on interpretable AI-driven hypothesis generation, ensuring that hypotheses are both transparent and actionable, making it particularly valuable in ethically sensitive fields such as governance and policy-making~\cite{shavit2023practices}. These tools demonstrate the power of interactive and collaborative systems to drive innovation by combining human expertise with AI-driven exploration.

The practical applications of interactive and collaborative systems span a wide array of disciplines, underscoring their transformative potential. In biomedical research, SciAgents has accelerated the identification of disease biomarkers and drug-target interactions by integrating expert feedback, contributing to faster development of diagnostic and therapeutic solutions~\cite{ghafarollahi2024sciagents}. In social sciences, CrowdScience has facilitated the collaborative generation of hypotheses on behavioral trends and social correlations, incorporating diverse stakeholder insights to uncover novel patterns~\cite{zhou2024hypothesisgenerationlargelanguage}. In ethics and governance, ExplanatoryAI has advanced policy development and regulatory frameworks by ensuring that hypotheses are interpretable and actionable, aiding decision-makers in evaluating the potential impacts of their choices~\cite{shavit2023practices}. These examples highlight the adaptability of interactive and collaborative systems in addressing domain-specific and interdisciplinary challenges.

The novelty of hypotheses generated by these systems lies in their ability to integrate human insight with AI-driven exploration. Human expertise provides contextual understanding, while AI expands the hypothesis space, enabling the discovery of innovative and practical solutions~\cite{zhou2024hypothesisgenerationlargelanguage}. However, reliance on human input can introduce biases or constrain the exploration of unconventional ideas~\cite{kim2018research}. To address this, recent advancements incorporate generative adversarial techniques (GANs) to balance human feedback with machine-generated alternatives, fostering a broader and more creative exploration of hypotheses~\cite{beltagy2019scibert}. These enhancements ensure that hypotheses are not only relevant but also push the boundaries of conventional thinking.

Interactive and collaborative systems represent a powerful paradigm for hypothesis generation, blending the creativity and expertise of humans with the computational efficiency and scalability of AI. By enabling iterative collaboration, fostering inclusivity, and ensuring transparency, these systems drive the creation of actionable and innovative hypotheses, paving the way for breakthroughs across diverse scientific fields.

\subsection{Causal Inference}

Causal inference frameworks~\cite{khatibi2024alcm, peters2017elements, lucas2007biomedical, neuberg2003causality} leverage advanced statistical and computational techniques to identify and analyze causal relationships within datasets. Unlike correlation-based approaches, which often fail to capture the underlying mechanisms of observed phenomena, causal inference frameworks focus on uncovering cause-and-effect pathways. This makes them particularly effective for generating actionable hypotheses grounded in causality, enabling researchers to propose interventions and predict their outcomes with greater confidence. By bridging the gap between observational data and experimental insights, causal inference frameworks have become indispensable tools across diverse scientific domains~\cite{jha2019hypothesis, qi2024large}.

At the core of causal inference methodologies are techniques designed to map, test, and validate causal relationships. Structural Causal Models (SCMs) represent one of the foundational approaches, using directed acyclic graphs (DAGs) to visualize and analyze causal pathways~\cite{jha2019hypothesis}. These models allow researchers to identify direct and indirect relationships between variables, providing a comprehensive framework for understanding complex systems. Bayesian networks complement SCMs by incorporating probabilistic reasoning to estimate the likelihood of causal links between variables~\cite{kim2018research}. This approach is particularly valuable in scenarios with uncertainty or incomplete information, where Bayesian reasoning helps prioritize plausible hypotheses. Interventional analysis takes causal inference a step further by simulating potential interventions within datasets to test causal hypotheses and predict the outcomes of specific actions~\cite{ghafarollahi2024sciagents}. Together, these techniques form a robust foundation for hypothesis generation, enabling researchers to derive mechanistic insights from observational data.

Several cutting-edge tools demonstrate the versatility and transformative potential of causal inference frameworks in hypothesis generation across diverse scientific domains. In biomedical research, CausalNet~\cite{peters2017elements}, which is based on Structural Causal Models (SCMs), maps complex causal pathways to generate hypotheses about disease progression and treatment effects, facilitating advances in understanding disease mechanisms and enabling predictive diagnostics and targeted therapies. In social and biomedical sciences, BayesCausality~\cite{lucas2007biomedical} leverages Bayesian networks to uncover causal relationships between socio-economic variables and health outcomes, providing robust, evidence-based insights for public health and policy interventions. In materials science, InterveneAI~\cite{neuberg2003causality} uses interventional analysis to simulate experimental conditions and identify causal relationships between material properties and performance metrics, accelerating the design and optimization of high-performance materials. Together, these tools illustrate how causal inference frameworks can generate mechanistically grounded hypotheses and support data-driven decision-making across biomedicine, social science, and engineering.

The novelty of hypotheses generated by causal inference frameworks lies in their ability to uncover mechanistic insights often overlooked by traditional methods. By focusing on causal relationships, these frameworks have identified unexpected treatment effects in biomedical datasets, challenging existing paradigms and opening new avenues for research~\cite{jha2019hypothesis}. However, the effectiveness of these frameworks is contingent on the quality of the input data. Incomplete datasets or the presence of latent variables can limit their capacity to generate innovative hypotheses, underscoring the importance of robust data collection and preprocessing~\cite{shavit2023practices}. To enhance novelty, researchers are increasingly integrating causal inference with machine learning techniques, such as deep generative models. This combination allows for the simultaneous capture of causal and non-causal patterns, broadening the scope of hypothesis generation while maintaining scientific rigor~\cite{beltagy2019scibert}.

Causal inference frameworks provide a powerful approach to understanding the underlying mechanisms of complex systems. By combining robust methodologies, advanced tools, and interdisciplinary applications, these frameworks enable researchers to generate actionable and novel hypotheses that drive scientific discovery and innovation. As the integration of causal inference with machine learning continues to evolve, the potential for these frameworks to redefine research paradigms and address pressing scientific challenges remains immense.

\subsection{Dynamic and Adaptive Knowledge Systems}

Dynamic and adaptive knowledge systems~\cite{fecho2021biomedical, yanscigraph, mosbach2020multiscale, rossi2020temporal}, particularly Dynamic Knowledge Graphs (DKGs), revolutionize hypothesis generation by incorporating temporal and real-time updates into traditional knowledge graph frameworks. Unlike static datasets, which can quickly become outdated, DKGs adapt to new data as it emerges, providing a flexible and responsive platform for capturing the evolving dynamics of scientific knowledge. This adaptability is critical in fast-paced fields such as biomedicine, environmental science, and chemical engineering, where maintaining the relevance and accuracy of hypotheses is paramount~\cite{kim2018research, qi2024large}.  

The methodologies underlying DKGs emphasize the integration of temporal, contextual, and real-time information to enhance hypothesis generation. Graph Temporal Networks (GTNs) serve as a foundational technique, encoding time-stamped edges to represent evolving relationships between entities. This enables researchers to uncover temporal patterns and shifts in data that static models might overlook. Real-time data integration dynamically updates graph structures, ensuring that hypotheses are continuously informed by the most current knowledge~\cite{gu2025forecastinghighimpactresearchtopics}. This capability is particularly valuable in disciplines where rapid discoveries demand immediate incorporation of new findings. Contextual embeddings further refine hypothesis precision by adapting graph representations to reflect the changing contexts of entities and their relationships, allowing for more nuanced and accurate insights. Together, these methodologies provide DKGs with the agility to address the dynamic needs of modern scientific inquiry. 

Several advanced tools demonstrate the transformative potential and interdisciplinary applicability of Dynamic Knowledge Graphs (DKGs) in scientific research. In biomedicine, UpdateKG~\cite{fecho2021biomedical} is a dynamic graph system that enables real-time tracking and refinement of gene-disease associations, supporting genomic research and personalized medicine by keeping hypotheses aligned with emerging discoveries. In environmental science, SciGraph~\cite{yanscigraph} dynamically models ecological interactions, such as predator-prey dynamics and environmental conditions, to generate actionable insights into ecosystem shifts and resilience. In chemical engineering, KG-Stream~\cite{mosbach2020multiscale} integrates streaming experimental data to predict chemical properties and refine hypotheses, accelerating the design and discovery of innovative materials and industrial processes. These tools highlight the adaptability of DKGs in addressing complex challenges across domains and underscore their value in generating timely, data-informed scientific hypotheses.

The novelty of hypotheses generated through DKGs lies in their ability to adapt to evolving datasets and uncover time-sensitive patterns. By leveraging time-aware updates, DKGs capture trends and insights that static systems often miss, such as transient gene-disease associations or short-lived environmental phenomena. However, reliance on real-time data streams introduces challenges, including the potential for incomplete or delayed updates that may limit the accuracy of generated hypotheses. To mitigate these limitations, integrating DKGs with predictive modeling techniques has emerged as a powerful enhancement. Predictive models enable DKGs to anticipate future trends, broadening the scope of hypothesis generation and fostering forward-looking insights.  

Dynamic and adaptive knowledge systems represent a significant advancement in the field of hypothesis generation. By seamlessly integrating temporal and contextual information with real-time updates, these systems provide researchers with a robust and flexible framework for exploring complex scientific questions. Their ability to generate hypotheses that are both innovative and actionable makes DKGs indispensable tools in the modern scientific landscape, driving progress across a diverse array of disciplines.

\subsection{Multi-Agent Systems}

Multi-agent systems (MAS)~\cite{su2024two, baek2024researchagent, kurakin2007unconventional, park2024leveraging} represent an innovative approach to hypothesis generation by deploying multiple autonomous agents that work collaboratively to explore complex hypothesis spaces. By assigning specialized roles to individual agents and fostering interactions among them, MAS enable the generation of hypotheses enriched by diverse expertise and cross-disciplinary insights. This collaborative dynamic mirrors the human scientific process, where teams with varied backgrounds and skills work together to address multifaceted challenges. MAS offer a scalable and efficient framework for hypothesis generation, particularly in domains requiring the integration of vast and heterogeneous datasets.  

The techniques underpinning MAS focus on role specialization, collaboration, and knowledge sharing to maximize the efficiency and creativity of hypothesis generation. Role-based specialization assigns distinct responsibilities to agents based on their domain expertise, ensuring that each agent focuses on a targeted aspect of the problem. This division of labor reduces redundancy and enhances the relevance of generated hypotheses. Negotiation and collaboration among agents enable the synthesis of partial hypotheses into cohesive and comprehensive propositions. By leveraging diverse perspectives, these interactions produce hypotheses that are more robust and interdisciplinary. Distributed knowledge sharing further accelerates the discovery process by integrating multi-domain insights, allowing agents to build upon each other’s findings and continuously refine their outputs. Together, these techniques make MAS a powerful tool for navigating complex scientific landscapes.

Several Multi-Agent System (MAS) frameworks exemplify the versatility and transformative potential of this methodology in hypothesis generation across scientific disciplines. In genomic research, TAIS (Team AI Scientists)~\cite{liu2024toward} enables agents to collaboratively analyze gene expression data and generate hypotheses on gene-protein interactions, advancing our understanding of molecular biology. In materials science, the MAS framework developed by~\cite{park2024leveraging} allows agents to explore chemical spaces and identify catalysts with unique properties, generating hypotheses that balance chemical feasibility with performance metrics and accelerating the discovery of high-performance materials. ResearchAgent~\cite{baek2024researchagent} simulates collaborative research environments in interdisciplinary settings, enabling the synthesis of insights from domains such as physics, chemistry, and biology to address complex, cross-domain scientific questions. A recent addition, VirSci~\cite{su2024two}, is a large-scale, LLM-based MAS designed to simulate real-world scientific collaboration. It assembles virtual scientist agents with distinct backgrounds to participate in structured discussions, novelty assessments, and iterative abstract refinement, showing significant gains in generating original research ideas compared to single-agent and prior multi-agent baselines. These frameworks collectively highlight MAS’s adaptability and capacity to drive innovation in modern scientific discovery.

The novelty of hypotheses generated by MAS lies in their ability to leverage diverse expertise and uncover multi-domain relationships. By integrating knowledge from different fields, MAS can propose groundbreaking hypotheses, such as linking gene expression patterns to disease phenotypes~\cite{kim2018research}. However, the novelty of these hypotheses can be constrained by communication and coordination overhead, which may hinder scalability and reduce the system's ability to address highly complex problems~\cite{zhou2024hypothesisgenerationlargelanguage}. To overcome these limitations, advanced negotiation protocols and enhanced knowledge-sharing mechanisms have been developed. These enhancements foster deeper interdisciplinary interactions, promoting the generation of more innovative and robust hypotheses~\cite{beltagy2019scibert}.  

Multi-agent systems represent a dynamic and collaborative framework for hypothesis generation, mirroring the diverse expertise and teamwork characteristic of human research teams. By integrating advanced techniques for collaboration and knowledge sharing, MAS has the potential to revolutionize scientific discovery, providing scalable and efficient solutions for addressing the complexities of modern research.  

\textbf{Conclusion.} The diverse approaches to hypothesis generation demonstrate the transformative potential of LLMs and computational techniques in advancing scientific discovery. Each method brings unique strengths: knowledge graphs uncover conceptual relationships, text mining highlights trends, machine learning drives novelty, retrieval-augmented generation enhances relevance, and multi-omics integration reveals insights into complex systems. Despite challenges like data constraints and interpretability, these methods promise to democratize hypothesis generation and accelerate breakthroughs across disciplines.  
Integrating these methodologies, such as combining machine learning with ontology-based reasoning or dynamic knowledge graphs with multi-agent systems, amplifies their strengths. Such hybrid strategies enable comprehensive and novel hypotheses, addressing complex problems and accelerating the pace of discovery in today’s evolving scientific landscape.  

\section{Categorization of Hypothesis Validation Approaches}
\label{sec:hypothesis_validation}

The rapid advancements in hypothesis generation systems, particularly those powered by LLMs and computational intelligence, have highlighted the need for rigorous validation frameworks. Unlike traditional hypothesis formulation, where human intuition and prior knowledge play a central role, AI-generated hypotheses require systematic evaluation to ensure they are novel, insightful, scientifically plausible, testable, and actionable. Effective validation mechanisms serve as a critical checkpoint in the research lifecycle, distinguishing viable hypotheses from speculative or erroneous propositions.

\begin{figure*}[ht] 
    \centering
    \includegraphics[width=\textwidth]{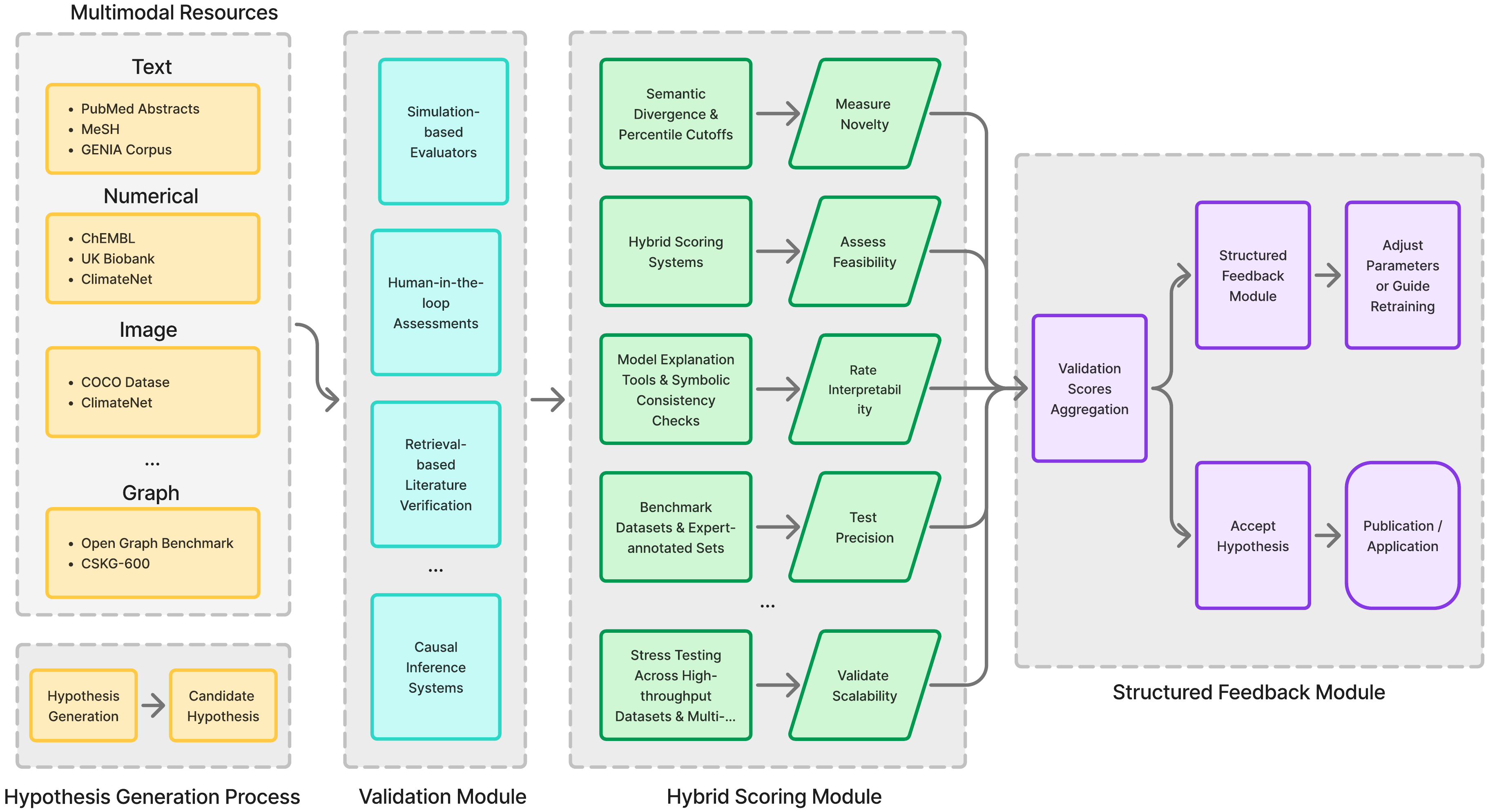}
    \caption{Pipeline for AI-assisted hypothesis validation. The figure outlines multiple validation modules, including simulation, human-in-the-loop assessments, retrieval-based verification, and causal inference, used to assess novelty, feasibility, interpretability, precision, and scalability. Finally, structured feedback and score aggregation are performed to guide acceptance, refinement, or retraining of hypotheses.}   
    \label{fig:hypothesis_generation_pipeline}
\end{figure*}

As scientific inquiries grow increasingly complex and interdisciplinary, traditional validation methodologies, such as empirical testing and statistical inference, have been augmented by simulation-driven analysis, predictive modeling, causal inference techniques, and collaborative validation systems. These modern approaches integrate real-time data streams, machine learning models, and multi-agent validation frameworks, offering more scalable, adaptive, and domain-specific verification strategies. The convergence of computational methods and empirical validation represents a paradigm shift in how hypotheses are evaluated across diverse scientific fields, including biomedicine, social sciences, materials science, and artificial intelligence research. Figure~\ref{fig:hypothesis_generation_pipeline} presents a multi-module architecture for hypothesis validation.

This section provides a structured taxonomy of hypothesis validation approaches, categorizing them based on their underlying methodologies, applications, and evaluation metrics. We explore how these approaches enhance reliability, reproducibility, and domain adaptability, ensuring that hypotheses meet the rigorous standards of contemporary scientific research. By examining these validation frameworks, we highlight the evolving role of AI-assisted validation in fostering credible and impactful discoveries across disciplines.

\begin{table}[ht]
\centering
\caption{Summary of Hypothesis Validation Approaches, Tools, and Metrics}
\resizebox{\columnwidth}{!}{%
\begin{tabular}{p{3cm}|p{2.5cm}|p{2cm}|p{3cm}|p{3cm}}
\hline
\textbf{Approach} & \textbf{Example Tool/System} & \textbf{Domain} & \textbf{Strengths} & \textbf{Metrics Used} \\ \hline
Experimental and Empirical Testing & LabKey Server~\cite{labkey} & Biomedical Research & Facilitates management and analysis of experimental data & Accuracy, Reproducibility Rate, Statistical Significance \\ \hline
Simulation-Driven Validation & Simulink~\cite{simulink} & Engineering, Robotics & Enables simulation of dynamic systems for hypothesis testing & Simulation Accuracy, Robustness Metric, Convergence Rate \\ \hline
Predictive, Adaptive, and Real-Time Validation & scikit-learn~\cite{scikit-learn}, Prophet~\cite{taylor2018prophet}, TensorFlow Serving~\cite{tensorflow-serving} & Machine Learning, AI Research & Uses predictive models, real-time adaptation, and statistical inference for evolving datasets & Prediction Accuracy, Model Fit (\(R^2\)), Latency, Adaptation Accuracy \\ \hline
Cross-Disciplinary Generalization & IBM Watson~\cite{ibm-watson} & Various Industries & Applies AI to ensure hypotheses are generalizable across domains & Transfer Learning Accuracy, Domain Alignment Score \\ \hline
Human-AI and Crowdsourced Validation & Zooniverse~\cite{zooniverse}, Jupyter Notebook~\cite{jupyter-notebook} & Citizen Science, AI, Education & Leverages human-AI interaction and crowdsourced expertise for validation & Agreement Rate, Consensus Score, User Satisfaction Metric, Iteration Success Rate \\ \hline
Causal Relationship Validation & TETRAD~\cite{tetrad} & Social Science, Medicine & Discovers causal relationships to validate hypotheses & Causal Effect Estimate, Goodness-of-Fit \\ \hline
Benchmarking and Standardized Testing & MLPerf~\cite{mlperf} & Machine Learning & Provides benchmarks for evaluating model performance & F1-Score, Comparative Improvement \\ \hline
Multi-Agent Validation & SciAgents~\cite{ghafarollahi2024sciagents} & AI, Computational Sciences & Leverages agent-based collaboration for scalable validation & Multi-Agent Consensus, Validation Confidence Score \\ \hline
Explainability and Interpretability Validation & SHAP~\cite{lundberg2017shap}, LIME~\cite{ribeiro2016lime} & AI, Social Sciences & Ensures interpretability of AI-generated hypotheses & Explainability Score, Feature Importance \\ \hline
\end{tabular}
}
\label{tab:validation_summary}
\end{table}

\subsection{Experimental and Empirical Testing}

Experimental validation remains a cornerstone of hypothesis testing, providing high reliability and precision through controlled experimentation. By manipulating variables and systematically observing their effects, this approach offers empirical evidence to confirm or refute hypotheses. It is particularly critical in fields such as biomedicine, chemistry, and physics, where direct causal inference is necessary to validate theoretical models~\cite{kim2018research}. Unlike computational approaches that rely on probabilistic modeling, experimental validation ensures direct empirical verification, making it indispensable for scientific rigor.

\textbf{Techniques and Tools.}  
Experimental validation employs several well-established methodologies to ensure accuracy, reproducibility, and generalizability. Controlled experiments minimize confounding factors by isolating specific variables, allowing precise measurement of their effects. Replication studies confirm the reliability of findings through repeated experimentation, ensuring consistency across different trials. Randomized Controlled Trials (RCTs) are a gold standard in biomedical research, enabling unbiased hypothesis testing by randomly assigning subjects to experimental conditions~\cite{beltagy2019scibert}. 

Integrating automation and AI-driven experimental platforms has significantly enhanced the efficiency of traditional workflows. LabKey Server is widely adopted in biomedical research, facilitating data management and real-time analysis for experimental validation~\cite{labkey}. In chemistry, robotic platforms integrated with AI-based automation improve reaction optimization and reduce human intervention, accelerating hypothesis testing~\cite{zhou2024hypothesisgenerationlargelanguage}. Automated microscopy systems in experimental biology enable real-time cellular imaging and validation, minimizing manual errors while enhancing experimental throughput~\cite{qi2024large}.

\textbf{Metrics for Validation.}  
To ensure robust evaluation of experimental outcomes, multiple quantitative metrics are employed. Accuracy measures the correctness of experimental results by determining the proportion of correctly identified true positives and true negatives. Precision assesses the consistency of results across repeated experiments. The reproducibility rate ensures that independent researchers can validate findings, thereby strengthening scientific credibility~\cite{sybrandt2018large}. Statistical significance, often determined through \( p \)-values (\( p < 0.05 \)), verifies that observed effects are unlikely due to random chance, further supporting hypothesis validation~\cite{beltagy2019scibert}.

\textbf{Practical Applications.}  
Experimental validation is pivotal across scientific disciplines, ensuring that hypotheses translate into reliable real-world insights. In chemistry, AI-driven robotic laboratories optimize reaction conditions for synthetic compound development, improving reaction efficiency and hypothesis testing~\cite{zhou2024hypothesisgenerationlargelanguage}. LabKey Server~\cite{labkey} supports large-scale validation of biomarker discoveries in biomedicine, contributing to advancements in precision medicine. In physics, experimental validation through particle accelerators facilitates empirical testing of subatomic particle interactions, providing direct evidence for theoretical frameworks~\cite{kim2018research}. These applications highlight the essential role of experimental validation in bridging the gap between theoretical research and empirical confirmation.

\textbf{Feasibility and Novelty Accessibility.}  
Despite its strengths, experimental validation is often resource-intensive and time-consuming, requiring significant investment in specialized equipment, controlled environments, and expert personnel. These constraints can limit scalability, particularly for large-scale hypothesis testing or research in resource-constrained settings~\cite{qi2024large}. Researchers increasingly integrate computational simulations and machine learning models with experimental validation to enhance feasibility, creating hybrid approaches that optimize resources while maintaining scientific rigor~\cite{zhou2024hypothesisgenerationlargelanguage}. 

Regarding novelty accessibility, experimental validation is a key enabler of breakthrough discoveries, offering empirical verification of innovative hypotheses. Unlike predictive models that rely on inferential reasoning, experimental validation provides direct observational evidence, ensuring that novel hypotheses are rigorously tested before broader adoption. However, the slow throughput of traditional experimental techniques may hinder the rapid exploration of unconventional hypotheses. Integrating automated high-throughput screening platforms and AI-assisted experimental workflows helps overcome this limitation, allowing researchers to efficiently test and validate a larger number of hypotheses~\cite{sybrandt2018large}.

Experimental validation remains essential to scientific discovery, ensuring that hypotheses are empirically tested and reproducible. With the integration of automation, AI-driven validation, and hybrid computational-experimental approaches, this methodology continues to evolve, addressing the growing complexities of modern scientific inquiry while maintaining scientific rigor and reliability.

\subsection{Simulation-Driven Validation}

Simulation-based validation employs advanced computational models to emulate real-world systems, offering a scalable and risk-free environment for hypothesis testing. This approach is particularly valuable for hypotheses that involve high costs, extended timeframes, or significant safety risks if tested through physical experiments. By iteratively modeling and analyzing complex systems in silico, simulations enable the precise refinement and validation of hypotheses, ensuring their feasibility and reliability before practical implementation~\cite{kim2018research}. 

A key strength of simulation-based validation lies in its ability to mitigate physical risks and reduce the resources required for experimentation. Controlled virtual environments allow researchers to test hypotheses without physical constraints, eliminating the need for expensive and time-consuming laboratory setups. For example, automotive crash simulations significantly reduce costs and safety risks compared to physical crash tests. Moreover, simulations scale effectively to accommodate large and multi-dimensional hypothesis spaces, enabling the simultaneous testing of numerous variables. This scalability and efficiency make simulation-driven validation an indispensable tool in modern scientific inquiry~\cite{beltagy2019scibert}.

\textbf{Techniques and Tools.}  
Simulation-based validation relies on several computational techniques to ensure robust hypothesis testing. Agent-Based Modeling (ABM) simulates the behavior of autonomous agents interacting in complex systems, making it highly applicable to ecology, economics, and social sciences~\cite{qi2024large}. Monte Carlo Simulations use stochastic sampling to assess probabilistic outcomes, enabling the evaluation of hypothesis robustness under uncertainty~\cite{kim2018research}. Finite Element Analysis (FEA) is widely used in engineering and materials science to model structural and mechanical properties, validating physical hypotheses related to stress distribution, material behavior, and mechanical failure~\cite{sybrandt2018large}. 

Several cutting-edge tools facilitate simulation-driven hypothesis validation. Simulink is a widely adopted platform for engineering and robotics, allowing researchers to model dynamic systems and test control strategies under simulated conditions~\cite{simulink}. BioSimulators, an advanced simulation toolkit for biomedical research, validates hypotheses related to molecular dynamics, enzyme kinetics, and cellular interactions~\cite{labkey}. CARLA, an open-source autonomous vehicle simulator, provides a virtual environment for testing hypotheses related to robotic navigation, traffic dynamics, and urban mobility~\cite{zhou2024hypothesisgenerationlargelanguage}. These tools exemplify the versatility and scalability of simulation-based validation, making it a powerful method for evaluating hypotheses across scientific disciplines.

\textbf{Metrics for Validation.}  
Simulation-based approaches employ several quantitative metrics to assess the validity, reliability, and robustness of simulated hypotheses. Simulation Accuracy measures how closely simulation outcomes align with empirical data, ensuring that the models effectively replicate real-world phenomena~\cite{beltagy2019scibert}. Convergence Rate evaluates how quickly simulations stabilize, reflecting the efficiency of iterative model refinement. Robustness Metrics assess how hypotheses perform across diverse simulated conditions, ensuring reliability under different parameter settings~\cite{kim2018research}. These metrics provide a comprehensive framework for evaluating the precision and effectiveness of simulation-based validation.

\textbf{Practical Applications.}  
Simulation-driven validation is widely applied across scientific and engineering domains, offering researchers a safe and scalable platform for hypothesis testing. In biomedicine, tools like BioSimulators validate hypotheses about drug interactions, enzyme activity, and disease progression, allowing researchers to refine models before clinical trials~\cite{labkey}. In robotics, CARLA enables the simulation of autonomous vehicle navigation, testing hypotheses related to sensor fusion, real-time decision-making, and safety protocols~\cite{zhou2024hypothesisgenerationlargelanguage}. In materials science, Finite Element Analysis (FEA) is employed to validate stress-strain predictions, ensuring that hypotheses regarding mechanical durability and performance hold under real-world conditions~\cite{sybrandt2018large}. These examples underscore the transformative role of simulation-based validation in hypothesis testing, providing risk-free and computationally scalable solutions.

\textbf{Feasibility and Novelty Accessibility.}  
Simulation-based validation is highly feasible, as it offers cost-effective and risk-free testing environments. Unlike laboratory-based experiments, simulations do not require physical resources, enabling researchers to iteratively refine models with minimal material costs~\cite{qi2024large}. However, the effectiveness of this method is contingent upon the accuracy and completeness of the underlying models. Simplifications and assumptions in simulations may introduce biases, potentially limiting their ability to capture real-world complexities~\cite{kim2018research}.

In terms of novelty accessibility, simulation-driven validation provides a safe and flexible environment for testing highly innovative and unconventional hypotheses. Unlike empirical testing, which requires physical prototypes and controlled conditions, simulations enable hypothesis exploration at scale, allowing researchers to test novel concepts that may not yet be feasible for real-world implementation~\cite{beltagy2019scibert}. Nevertheless, the interpretability of simulation results remains a challenge, as complex computational models often require post-validation through empirical experiments to ensure external validity~\cite{sybrandt2018large}.

Simulation-driven validation continues to evolve as an indispensable methodology in modern scientific research. By integrating advanced computational techniques, AI-driven automation, and real-time predictive modeling, simulation-based validation offers a scalable, efficient, and innovative framework for hypothesis testing. As computational models grow increasingly sophisticated, their role in scientific discovery and technological innovation will continue to expand, ensuring that simulation-driven validation remains a cornerstone of modern hypothesis evaluation.

\subsection{Predictive, Adaptive, and Real-Time Validation}

Predictive, adaptive, and real-time validation leverages machine learning models, statistical inference, and dynamic adaptation to evaluate hypotheses in a scalable, cost-effective, and data-driven manner. Unlike experimental approaches that require physical testing, these methods validate hypotheses through predictive modeling, real-time adjustments, and adaptive learning mechanisms. These approaches are particularly valuable for large-scale, complex, or continuously evolving datasets, making them indispensable in biomedicine, climate modeling, AI research, and financial forecasting~\cite{bengio2019meta}.

A defining characteristic of predictive and adaptive validation is its ability to integrate high-dimensional datasets and continuous feedback loops, refining hypotheses and models iteratively. By enabling real-time hypothesis evaluation, these approaches dynamically adjust predictions as new data emerges, ensuring up-to-date and context-aware validation. This adaptability is crucial in fields where real-world conditions change rapidly, such as autonomous systems, financial markets, and environmental monitoring~\cite{beltagy2019scibert}.

\textbf{Techniques and Tools.}  
Predictive validation encompasses multiple modeling techniques to ensure robust hypothesis evaluation. Supervised learning models, widely applied in biomedicine and AI research, rely on labeled datasets for pattern recognition and classification~\cite{qi2024large}. Bayesian inference models integrate prior knowledge with observed data, facilitating probabilistic hypothesis validation and dynamic uncertainty estimation, particularly in social sciences and healthcare analytics~\cite{kim2018research}. Time-series forecasting models, commonly used in climate science and financial analysis, predict trends and validate hypotheses based on historical and real-time data~\cite{zhou2024hypothesisgenerationlargelanguage}.

Several tools exemplify the application of these techniques. Scikit-learn provides robust validation frameworks for predictive modeling across various disciplines~\cite{scikit-learn}. Prophet, a forecasting tool developed by Meta, specializes in time-series prediction and anomaly detection, aiding real-time hypothesis validation in economics and environmental sciences~\cite{taylor2018prophet}. TensorFlow Serving enables adaptive hypothesis validation in AI models, allowing real-time inference updates as new data becomes available~\cite{tensorflow-serving}. These tools highlight the versatility of predictive and adaptive validation, ensuring that hypotheses evolve dynamically with incoming data streams.

\textbf{Metrics for Validation.}  
To ensure rigorous evaluation, predictive validation relies on well-defined quantitative metrics. Prediction Accuracy assesses the correctness of model predictions by comparing forecasted outcomes with observed data~\cite{beltagy2019scibert}. Bayesian Posterior Probability evaluates the likelihood of a hypothesis given available evidence, integrating both prior knowledge and new observations~\cite{kim2018research}. Model Fit (\(R^2\)) measures how well predictive models explain the variance in real-world datasets, serving as a key indicator of reliability~\cite{scikit-learn}. Additionally, Adaptation Accuracy evaluates how effectively real-time models adjust to evolving data streams, ensuring continuous model refinement and reliability~\cite{bengio2019meta}.

\textbf{Practical Applications.}  
Predictive, adaptive, and real-time validation plays a crucial role across scientific domains, enabling scalable and efficient hypothesis testing. In genomics, predictive models identify gene-disease associations, accelerating hypothesis validation in personalized medicine~\cite{qi2024large}. In climate science, time-series forecasting models, such as Prophet, predict long-term temperature variations, aiding the validation of hypotheses about climate change and environmental feedback loops~\cite{taylor2018prophet}. In AI research, TensorFlow Serving supports adaptive model validation, ensuring that machine learning models remain accurate as they process real-time streaming data~\cite{tensorflow-serving}. These applications underscore the significance of predictive validation in handling complex, dynamic, and continuously evolving scientific challenges.

\textbf{Feasibility and Novelty Accessibility.}  
Predictive validation offers high feasibility due to its cost-effectiveness and scalability. Unlike traditional experimental validation, which requires physical setups and manual intervention, predictive approaches allow researchers to test and refine hypotheses iteratively using existing datasets and real-time observations. However, the effectiveness of predictive validation depends on the availability of high-quality training data and the interpretability of complex models~\cite{zhou2024hypothesisgenerationlargelanguage}. Poorly curated datasets or biased training samples can lead to overfitting, reducing the generalizability of validated hypotheses~\cite{beltagy2019scibert}.

In terms of novelty accessibility, predictive validation enables hypothesis exploration in high-dimensional spaces, revealing patterns and relationships that might be overlooked in traditional experimental settings. This approach is particularly valuable in AI-assisted discovery, where deep learning models generate hypotheses that human researchers might not have considered~\cite{sybrandt2018large}. However, the insights generated by predictive models are often constrained by the assumptions inherent in the models themselves, limiting their ability to fully explore unconventional or outlier hypotheses. As a result, hybrid validation approaches—combining predictive modeling with empirical validation—are increasingly being adopted to enhance both scalability and interpretability.

Predictive, adaptive, and real-time validation continues to evolve as an essential methodology for hypothesis testing, bridging the gap between static theoretical models and dynamic real-world applications. As advancements in machine learning, real-time analytics, and automated hypothesis refinement progress, predictive validation will play an increasingly critical role in accelerating scientific discovery and technological innovation.

\subsection{Cross-Disciplinary Generalization}

Cross-domain validation is a powerful approach for assessing hypotheses by evaluating their applicability across multiple scientific fields or datasets. This methodology ensures that hypotheses are generalizable and robust, making it particularly valuable in interdisciplinary research where knowledge transfer and cross-disciplinary correlations are essential. By testing hypotheses in diverse domains, cross-domain validation facilitates the discovery of broader applications and fosters the integration of knowledge across scientific disciplines.

A key strength of cross-domain validation lies in its ability to encourage generalization by testing hypotheses across varying fields. This approach not only ensures the relevance of hypotheses beyond their original context but also enables the discovery of novel insights that span multiple domains. Additionally, cross-domain validation facilitates the transfer of knowledge and methodologies, providing a framework for integrating diverse datasets and experimental techniques, thereby strengthening the coherence and applicability of hypotheses.

\textbf{Techniques and Tools.}  
The effectiveness of cross-domain validation is supported by several innovative techniques. \textit{Domain Mapping} aligns datasets and methodologies across fields to ensure consistency and compatibility during validation, addressing challenges of heterogeneity in cross-domain data~\cite{zhang2024decoupled}. \textit{Interdisciplinary Networks} utilize network models to identify shared structures and relationships, enhancing the coherence of hypotheses by uncovering commonalities across scientific disciplines~\cite{sybrandt2018large}. \textit{Transfer Learning} is another critical technique, applying knowledge derived from one domain to test and validate hypotheses in another, enabling adaptability and scalability in hypothesis evaluation~\cite{wu2020comprehensive}.

Several tools and systems exemplify the application of these techniques. \textit{CrossValNet} integrates datasets from genomics, pharmacology, and environmental science, enabling researchers to validate hypotheses that span these fields~\cite{sybrandt2018large}. \textit{InterdisciplinaryTest} facilitates hypothesis validation across domains such as physics, chemistry, and biology, ensuring compatibility and coherence in results~\cite{zhou2024hypothesisgenerationlargelanguage}. \textit{TransferTest} employs transfer learning techniques to adapt and validate AI models and hypotheses across diverse scientific domains, highlighting its versatility and impact~\cite{touvron2023llama}.

\textbf{Metrics for Validation.}  
Cross-domain validation employs several metrics to assess the robustness and applicability of hypotheses. \textit{Domain Alignment Score} measures the consistency of results across different domains, providing a quantitative evaluation of hypothesis generalizability. \textit{Transfer Learning Accuracy} evaluates the effectiveness of knowledge transfer during validation, ensuring that models and hypotheses perform reliably in target domains. \textit{Interdisciplinary Correlation Metric} assesses the strength of relationships between variables in different fields, enabling researchers to identify and quantify cross-disciplinary connections. These metrics collectively provide a comprehensive framework for evaluating the success of cross-domain validation efforts.

\textbf{Practical Applications.}  
The practical applications of cross-domain validation span a wide range of scientific fields, underscoring its adaptability and utility. In genomics and pharmacology, \textit{CrossValNet} has validated gene-disease associations by integrating genomic and pharmacological datasets, demonstrating the broader relevance of these hypotheses~\cite{sybrandt2018large}. In environmental science and biology, \textit{InterdisciplinaryTest} has explored connections between ecological variables and biological diversity, providing actionable insights into conservation strategies~\cite{zhou2024hypothesisgenerationlargelanguage}. In AI and robotics, \textit{TransferTest} has validated hypotheses about navigation algorithms by transferring models from robotics to autonomous vehicles, showcasing its ability to ensure adaptability across diverse scientific fields~\cite{touvron2023llama}.

\textbf{Feasibility and Novelty Accessibility.}  
Cross-domain validation is highly feasible for fostering hypothesis robustness and facilitating the integration of diverse datasets and methodologies. Its strengths lie in its capacity to generalize hypotheses across multiple domains, ensuring their reliability and applicability. However, the approach requires extensive expertise across different fields and significant efforts to align heterogeneous datasets, which can pose challenges.

In terms of novelty, cross-domain validation excels in facilitating the discovery of novel connections and insights by leveraging domain-specific knowledge. By integrating datasets and methodologies from diverse disciplines, this approach uncovers relationships that might otherwise remain hidden. However, the availability and compatibility of cross-domain datasets may limit its broader applications, particularly in emerging or underexplored fields.

Cross-domain validation represents a transformative methodology for hypothesis testing in interdisciplinary research. By encouraging generalization, uncovering cross-disciplinary correlations, and fostering knowledge transfer, this approach continues to drive innovation and discovery across diverse scientific landscapes.

\subsection{Human-AI and Crowdsourced Validation}

Human-AI and crowdsourced validation leverage both computational efficiency and collective intelligence to test and evaluate hypotheses. This combined approach democratizes hypothesis validation by integrating human intuition, domain expertise, and machine learning capabilities, enabling large-scale, iterative, and context-aware evaluations. Crowdsourced validation engages diverse participant groups to provide broad insights, while Human-AI collaboration refines hypotheses through interactive feedback loops and explainable AI, enhancing interpretability and robustness. These methodologies are particularly beneficial in domains requiring broad engagement, adaptability, and transparency.

A defining strength of this approach is its ability to combine the scalability of crowdsourced validation with the precision and adaptability of Human-AI collaboration. Crowdsourced validation ensures that hypotheses are tested across a wide range of perspectives, while AI-driven refinements optimize accuracy and efficiency. Furthermore, iterative feedback loops foster continuous hypothesis improvement, ensuring that both human insights and machine learning capabilities contribute dynamically to the validation process.

\textbf{Techniques and Tools.}  
The effectiveness of Human-AI and crowdsourced validation is supported by several innovative techniques. \textit{Open Feedback Systems} collect real-time user feedback, enabling dynamic adjustments during hypothesis validation~\cite{zhou2024hypothesisgenerationlargelanguage}. \textit{Gamification} engages participants through interactive elements to sustain engagement and improve data quality~\cite{kim2018research}. \textit{Consensus Aggregation} synthesizes diverse participant inputs to derive statistically robust conclusions, mitigating outlier influence and ensuring reliability~\cite{sybrandt2018large}. Additionally, \textit{Interactive AI Feedback} facilitates iterative refinements by allowing users to guide AI-driven hypothesis improvements, ensuring adaptability and contextual relevance~\cite{zhou2024hypothesisgenerationlargelanguage}.

Several tools exemplify the application of these techniques. \textit{Zooniverse} provides a crowdsourced platform where participants contribute to scientific discovery by validating hypotheses across various disciplines~\cite{zooniverse}. \textit{Jupyter Notebook} facilitates collaborative hypothesis testing by integrating human expertise with AI-driven analytics~\cite{jupyter-notebook}. \textit{ExplainValid} combines explainable AI techniques with user feedback, improving transparency and trust in hypothesis validation processes~\cite{zhou2024hypothesisgenerationlargelanguage}. \textit{FeedbackLoopAI} integrates iterative user feedback into machine learning models, ensuring that hypotheses evolve dynamically based on human insights~\cite{sybrandt2018large}.

\textbf{Metrics for Validation.}  
To ensure the reliability and effectiveness of Human-AI and crowdsourced validation, multiple quantitative metrics are employed. The \textit{Agreement Rate} measures response consistency across participants, providing insight into hypothesis reliability. The \textit{Consensus Score} quantifies the degree of collective agreement, ensuring robust validation through diverse perspectives. The \textit{User Engagement Metric} evaluates the level of participant involvement, indicating the scalability and accessibility of the validation process. In Human-AI interactions, the \textit{Iteration Success Rate} assesses how effectively hypotheses are refined over successive validation rounds, while the \textit{Explainability Score} quantifies the interpretability of AI-generated outputs based on human verification~\cite{zhou2024hypothesisgenerationlargelanguage}.

\textbf{Practical Applications.}  
The integration of Human-AI and crowdsourced validation has demonstrated significant utility across scientific fields. In social sciences, \textit{Zooniverse} has validated hypotheses about public perception and social behaviors through large-scale participant contributions~\cite{zooniverse}. In Human-Computer Interaction (HCI), \textit{Jupyter Notebook} has facilitated usability testing by integrating human insights with machine learning-driven analytics~\cite{jupyter-notebook}. In AI ethics, \textit{ExplainValid} has enabled researchers to validate ethical guidelines by offering transparent, interpretable AI-generated recommendations~\cite{zhou2024hypothesisgenerationlargelanguage}. Additionally, in healthcare, \textit{FeedbackLoopAI} has assisted clinicians in refining diagnostic models through iterative AI-assisted hypothesis validation~\cite{sybrandt2018large}.

\textbf{Feasibility and Novelty Accessibility.}  
Human-AI and crowdsourced validation offer high feasibility due to their scalability, cost-effectiveness, and accessibility. Crowdsourced validation provides diverse perspectives while minimizing resource constraints, making it particularly useful for hypothesis testing in large and distributed participant groups. However, challenges such as demographic biases and response variability must be managed to ensure generalizability. In contrast, Human-AI collaboration enhances feasibility in complex, domain-specific tasks requiring expert intervention, though it may be limited by scalability constraints.

In terms of novelty, this hybrid approach excels at uncovering unconventional hypotheses through broad participant contributions and AI-driven insights. Crowdsourced validation fosters creativity by integrating diverse viewpoints, while Human-AI collaboration enhances the precision and interpretability of results. However, reliance on human input can introduce biases, and AI models may struggle with outlier hypotheses that deviate from learned patterns. Addressing these limitations requires careful design of validation frameworks that optimize both human intuition and machine-driven analysis.

Human-AI and crowdsourced validation represent a transformative approach to hypothesis testing, combining the power of collective intelligence with computational efficiency. By leveraging iterative feedback, interdisciplinary collaboration, and scalable participation, this approach fosters inclusivity, adaptability, and rigor across a wide range of scientific and practical applications.

\subsection{Causal Relationship Validation}

Causal relationship validation is a critical methodology for establishing cause-and-effect relationships between variables, distinguishing genuine causal mechanisms from mere correlations. This approach is particularly valuable in fields such as social sciences, biomedicine, and economics, where controlled experiments may be impractical or infeasible. By leveraging both observational and experimental data, causal inference techniques provide a structured framework for hypothesis validation, allowing for deeper insights into underlying mechanisms.

A key strength of causal relationship validation lies in its focus on causality rather than associative patterns, ensuring that validated hypotheses have mechanistic explanations beyond statistical relationships. This methodological rigor enhances the applicability of hypotheses in real-world decision-making processes, making it essential in fields requiring strong evidence-based conclusions. Additionally, causal inference techniques bridge the gap between observational studies and actionable insights, reinforcing their importance in modern scientific research.

\textbf{Techniques and Tools.}  
Causal inference validation employs several advanced methodologies to establish reliable causal relationships. \textit{Structural Causal Models (SCMs)} use directed acyclic graphs (DAGs) to explicitly represent causal pathways, enabling systematic validation of mechanistic hypotheses~\cite{pearl2009causal}. \textit{Propensity Score Matching (PSM)} reduces confounding effects by simulating randomization in observational datasets, improving the robustness of causal conclusions~\cite{rosenbaum1983central}. \textit{Instrumental Variable Analysis (IVA)} introduces external instruments to estimate causal effects when direct experimentation is infeasible, making it particularly useful in econometric studies~\cite{angrist1996identification}.

Several computational tools facilitate the practical implementation of these techniques. \textit{TETRAD}, a widely used causal discovery platform, applies SCMs to infer causal structures in domains such as epidemiology and behavioral sciences~\cite{tetrad}. \textit{Scikit-learn}, though primarily a machine learning library, includes causal inference modules that support propensity score-based methods for causal validation~\cite{scikit-learn}. \textit{IBM Watson} employs AI-driven causal modeling techniques to validate hypotheses across multiple industries, offering insights into customer behavior, healthcare diagnostics, and economic forecasting~\cite{ibm-watson}.

\textbf{Metrics for Validation.}  
Causal inference validation employs several quantitative metrics to assess the strength and reliability of causal relationships. The \textit{Causal Effect Estimate} quantifies the magnitude of causal influences by comparing outcomes across treated and control groups, ensuring precise hypothesis evaluation~\cite{rosenbaum1983central}. The \textit{Goodness-of-Fit for SCMs} evaluates how well causal models explain observed data, offering a quantitative measure of model reliability~\cite{pearl2009causal}. Additionally, the \textit{Instrument Relevance Score} assesses the validity of instrumental variables, ensuring that they provide unbiased causal estimates and meet key econometric assumptions~\cite{angrist1996identification}.

\textbf{Practical Applications.}  
Causal inference validation has been successfully applied across a range of disciplines. In biomedical research, \textit{TETRAD} has been instrumental in identifying causal links between genetic variants and disease phenotypes, advancing precision medicine~\cite{tetrad}. In policy evaluation, \textit{IBM Watson} has been used to assess the causal impact of education and healthcare policies, aiding policymakers in evidence-based decision-making~\cite{ibm-watson}. In econometrics, \textit{Scikit-learn}'s causal inference techniques have facilitated the evaluation of fiscal policies and labor market trends, supporting data-driven economic strategies~\cite{scikit-learn}.

\textbf{Feasibility and Novelty Accessibility.}  
Causal relationship validation offers high feasibility due to its adaptability to both observational and experimental data. Its ability to infer causality from complex datasets makes it a valuable tool across multiple domains. However, challenges such as confounding bias, instrument validity, and model specification errors must be carefully managed to ensure accurate results. The effectiveness of causal inference methods depends on the availability of high-quality data and the correct selection of causal assumptions, which can be limiting factors in some studies.

In terms of novelty, causal inference validation excels at uncovering previously unidentified causal pathways, enriching scientific understanding. By providing mechanistic explanations, it extends hypothesis validation beyond surface-level correlations, enabling deeper insights into fundamental processes. However, causal inference models rely on assumptions such as the validity of instrumental variables and the correctness of DAG structures, which may constrain the exploration of novel hypotheses. Advances in hybrid causal discovery techniques and the integration of AI-driven causal modeling promise to further enhance the scope and accessibility of causal inference validation.

Causal relationship validation remains a cornerstone of modern scientific research, offering robust methodologies to uncover the fundamental drivers of complex phenomena. As computational techniques and data integration frameworks continue to evolve, causal inference will play an increasingly critical role in refining hypotheses and guiding evidence-based decision-making across diverse scientific disciplines.

\subsection{Benchmarking and Standardized Testing}

Benchmarking and standardized testing serve as foundational approaches for hypothesis evaluation, utilizing predefined datasets, protocols, and metrics to ensure reliability, accuracy, and comparability. This methodology is particularly prevalent in fields such as machine learning, artificial intelligence, and experimental sciences, where the ability to systematically compare hypotheses against established baselines fosters reproducibility and facilitates objective assessment. By providing quantifiable measures of performance, benchmarking enables researchers to identify strengths, weaknesses, and areas for improvement in their hypotheses, accelerating scientific and technological advancements.

A key characteristic of benchmarking is its emphasis on reproducibility and consistency, ensuring that hypotheses can be evaluated across diverse research environments. Standardized testing offers a structured framework for assessing performance relative to existing solutions, enabling transparent and evidence-based decision-making. This comparative approach also helps identify the competitive advantages of new hypotheses, facilitating broader adoption in real-world applications.

\textbf{Techniques and Tools.}  
Benchmarking and standardized testing employ a range of established techniques to ensure rigorous validation. \textit{Baseline Comparison} assesses hypothesis performance against predefined reference models, providing a direct measure of improvement~\cite{mlperf}. \textit{Cross-Validation}, widely used in machine learning and experimental sciences, partitions data into training and testing subsets to ensure robustness and generalizability~\cite{scikit-learn}. \textit{Performance Metrics Analysis} applies standardized quantitative measures, such as accuracy, F1-score, and comparative improvement, to comprehensively evaluate hypotheses~\cite{mlperf}.

Several tools exemplify these techniques in various domains. \textit{MLPerf}, an industry-standard benchmarking suite, provides reproducible evaluations of machine learning models across multiple tasks, enabling rigorous hypothesis testing~\cite{mlperf}. \textit{Scikit-learn}, a widely used machine learning library, includes built-in benchmarking utilities for evaluating model generalization across datasets~\cite{scikit-learn}. \textit{Simulink}, a simulation platform, facilitates the comparative testing of control systems and engineering hypotheses under standardized experimental conditions~\cite{simulink}.

\textbf{Metrics for Validation.}  
Standardized testing employs well-defined metrics to assess hypothesis performance objectively. \textit{Accuracy} measures correctness by comparing predicted outcomes with observed results:
\[
\text{Accuracy} = \frac{\text{Correct Predictions}}{\text{Total Predictions}}
\]
The \textit{F1-Score}, balancing precision and recall, is particularly useful for hypotheses evaluated on imbalanced datasets:
\[
F1 = 2 \cdot \frac{\text{Precision} \cdot \text{Recall}}{\text{Precision} + \text{Recall}}
\]
Additionally, \textit{Comparative Improvement} quantifies the relative performance gain of a hypothesis over baseline solutions:
\[
\text{Improvement} = \frac{\text{Hypothesis Performance} - \text{Baseline Performance}}{\text{Baseline Performance}}
\]

\textbf{Practical Applications.}  
Benchmarking and standardized testing have demonstrated their utility across a broad range of disciplines. In machine learning, \textit{MLPerf} has validated neural network architectures by benchmarking their performance on datasets such as ImageNet, driving advancements in model optimization and evaluation~\cite{mlperf}. In biomedical science, \textit{Scikit-learn} has supported hypothesis validation in genomics and disease prediction, ensuring reproducibility in clinical applications~\cite{scikit-learn}. In engineering and robotics, \textit{Simulink} has benchmarked control algorithms and navigation models under standardized experimental conditions, facilitating the development of reliable autonomous systems~\cite{simulink}.

\textbf{Feasibility and Novelty Accessibility.}  
The feasibility of benchmarking lies in its structured and reproducible validation framework, which accelerates the adoption of validated hypotheses by quantifying their advantages over existing solutions. However, its effectiveness depends on the availability and relevance of benchmark datasets, which may become outdated or fail to capture emerging trends, limiting their applicability.

In terms of novelty, benchmarking effectively highlights innovations by measuring hypotheses relative to established baselines. This ensures that advancements are grounded in objective comparisons and reproducible evidence. However, reliance on predefined benchmarks can sometimes constrain exploratory research, as benchmarks may favor incremental improvements over paradigm-shifting hypotheses.

Benchmarking and standardized testing provide a rigorous and structured methodology for hypothesis evaluation, ensuring reliability, comparability, and scientific rigor. By leveraging standardized datasets and performance metrics, this approach fosters reproducibility and drives progress across machine learning, experimental sciences, and engineering disciplines.

\subsection{Multi-Agent Validation}

Multi-agent validation leverages autonomous and collaborative agents to systematically test and refine hypotheses, providing a scalable and distributed approach to validation. By utilizing multiple intelligent agents that interact, exchange information, and adapt their strategies, this methodology enables efficient validation in complex, dynamic, and data-rich environments. Multi-agent validation is particularly valuable in artificial intelligence, computational sciences, and decision-making systems, where multiple perspectives and decentralized reasoning enhance the reliability and robustness of validated hypotheses.

A defining characteristic of multi-agent validation is its ability to simulate complex real-world interactions by distributing hypothesis evaluation across multiple agents. These agents operate in parallel, facilitating large-scale hypothesis validation while dynamically adjusting their strategies based on continuous feedback. This approach enhances robustness by reducing bias and increasing the diversity of perspectives incorporated into the validation process.

\textbf{Techniques and Tools.}  
Multi-agent validation employs a range of techniques to coordinate and optimize the validation process. \textit{Consensus Mechanisms} ensure that agents reach agreement on hypothesis validity through iterative feedback and refinement, enhancing reliability~\cite{ghafarollahi2024sciagents}. \textit{Distributed Learning} enables agents to learn and adapt to new data independently while collectively improving hypothesis validation outcomes~\cite{ghafarollahi2024sciagents}. \textit{Reinforcement Learning-Based Coordination} optimizes agent collaboration, allowing them to efficiently explore and validate hypotheses in complex environments~\cite{ghafarollahi2024sciagents}.

Several computational frameworks support multi-agent validation across different domains. \textit{SciAgents}, a multi-agent system designed for hypothesis generation and validation, coordinates autonomous agents to evaluate scientific claims with high efficiency~\cite{ghafarollahi2024sciagents}. \textit{Protagents} integrates multiple AI-driven agents to validate hypotheses in dynamic knowledge environments, improving adaptability and inference capabilities~\cite{ghafarollahi2024protagents}. \textit{Agent-based Decision Platforms} utilize decentralized agents for hypothesis testing in strategic decision-making and simulation-driven research, allowing for scalable and real-time validation~\cite{ghafarollahi2024sciagents}.

\textbf{Metrics for Validation.}  
Multi-agent validation relies on specialized metrics to assess hypothesis consistency and agent performance. The \textit{Multi-Agent Consensus Score} quantifies the degree of agreement among agents, ensuring reliability in hypothesis evaluation:
\[
\text{Consensus Score} = \frac{\text{Number of Agents in Agreement}}{\text{Total Number of Agents}}
\]
The \textit{Validation Confidence Score} evaluates the confidence level of hypothesis acceptance, incorporating uncertainty estimates from multiple agents:
\[
\text{Validation Confidence} = \frac{\sum \text{Agent Confidence Scores}}{\text{Number of Agents}}
\]
Additionally, the \textit{Exploration-Exploitation Balance} measures how well agents optimize hypothesis validation by balancing novel exploration with confirmatory validation:
\[
\text{Exploration-Exploitation Ratio} = \frac{\text{Exploratory Actions}}{\text{Confirmatory Actions}}
\]

\textbf{Practical Applications.}  
Multi-agent validation has demonstrated its effectiveness across diverse scientific and technological domains. In AI research, \textit{SciAgents} has been used to validate machine learning-generated hypotheses by leveraging collaborative AI agents, ensuring improved generalization and robustness~\cite{ghafarollahi2024sciagents}. In scientific discovery, \textit{Protagents} has facilitated hypothesis validation by enabling autonomous knowledge exploration in interdisciplinary research~\cite{ghafarollahi2024protagents}. In computational sciences, multi-agent decision platforms have been applied to strategic planning and simulation-based hypothesis validation, improving decision accuracy in complex systems~\cite{ghafarollahi2024sciagents}.

\textbf{Feasibility and Novelty Accessibility.}  
Multi-agent validation offers high feasibility due to its ability to scale hypothesis validation processes efficiently. By distributing computational tasks among agents, this approach enables large-scale validation without requiring centralized control. However, challenges such as agent coordination, communication overhead, and decision-making biases must be carefully managed to maintain validation accuracy.

In terms of novelty, multi-agent validation fosters innovative hypothesis exploration by leveraging independent but interconnected agents. This decentralized approach allows for the discovery of unconventional insights that may be overlooked in traditional validation frameworks. However, the reliance on multi-agent interactions introduces complexity, requiring robust mechanisms to manage conflicts, align validation strategies, and ensure interpretability.

Multi-agent validation continues to revolutionize hypothesis testing by integrating autonomous decision-making, distributed learning, and collaborative inference. As AI and computational sciences advance, multi-agent systems will play an increasingly critical role in ensuring scalable, efficient, and high-confidence hypothesis validation across scientific disciplines.

\subsection{Explainability and Interpretability Validation}

Explainability and interpretability validation ensures that hypotheses and their underlying models are not only accurate but also transparent and understandable. This approach is particularly crucial in domains such as artificial intelligence, social sciences, and biomedical research, where the ability to explain results fosters trust, reproducibility, and informed decision-making. By validating hypotheses through interpretable frameworks, this methodology enhances transparency, accountability, and the reliability of complex models.

A defining characteristic of explainability and interpretability validation is its focus on making complex hypothesis evaluation processes understandable to both domain experts and non-specialists. This is particularly important in AI-driven hypothesis generation, where black-box models can produce highly accurate but opaque results. Explainability validation helps bridge this gap by ensuring that the reasoning behind hypothesis acceptance or rejection is accessible and interpretable.

\textbf{Techniques and Tools.}  
Explainability and interpretability validation employ various techniques to clarify how hypotheses are derived and assessed. \textit{Feature Attribution Methods} identify the most influential variables in hypothesis evaluation, ensuring transparency in model-driven hypothesis testing~\cite{lundberg2017shap}. \textit{Local Interpretable Model-Agnostic Explanations (LIME)} approximates black-box models with simpler, interpretable models to provide local explanations for hypothesis validation~\cite{ribeiro2016lime}. \textit{Counterfactual Reasoning} generates "what-if" scenarios to test how slight modifications in input variables affect the hypothesis outcome, ensuring robustness and fairness~\cite{ribeiro2016lime}.

Several tools facilitate explainability and interpretability validation. \textit{SHAP (Shapley Additive Explanations)} provides game-theoretic explanations for complex models, quantifying feature importance in hypothesis validation~\cite{lundberg2017shap}. \textit{LIME} generates local surrogate models to make AI-generated hypotheses more interpretable~\cite{ribeiro2016lime}. \textit{IBM Watson Explainability Tools} integrate explainability methods into AI-driven hypothesis testing, ensuring that predictions align with human reasoning and regulatory requirements~\cite{ibm-watson}.

\textbf{Metrics for Validation.}  
Explainability validation relies on well-defined metrics to assess interpretability and transparency. The \textit{Explainability Score} quantifies the degree to which a hypothesis or model outcome can be understood by human users:
\[
\text{Explainability Score} = \frac{\text{Number of Correctly Interpreted Predictions}}{\text{Total Predictions}}
\]
The \textit{Feature Importance Consistency} metric evaluates how consistently a model ranks the importance of different variables across different runs:
\[
\text{Feature Consistency} = \frac{\text{Stable Feature Rankings across Runs}}{\text{Total Features Ranked}}
\]
Additionally, the \textit{Human Trust Index} measures user confidence in hypothesis validation results based on their clarity and interpretability:
\[
\text{Trust Index} = \frac{\text{Number of Users Who Trust Explanations}}{\text{Total Users Surveyed}}
\]

\textbf{Practical Applications.}  
Explainability and interpretability validation have demonstrated significant impact across diverse domains. In AI research, \textit{SHAP} has been applied to validate machine learning hypotheses by explaining the influence of different features on model predictions, improving model trustworthiness~\cite{lundberg2017shap}. In biomedical science, \textit{LIME} has been used to interpret AI-driven disease diagnosis hypotheses, ensuring that medical predictions align with clinical reasoning~\cite{ribeiro2016lime}. In regulatory compliance, \textit{IBM Watson Explainability Tools} have facilitated transparent hypothesis validation in financial and legal decision-making, ensuring adherence to explainability mandates~\cite{ibm-watson}.

\textbf{Feasibility and Novelty Accessibility.}  
Explainability validation is increasingly feasible due to its widespread adoption in fields such as artificial intelligence, healthcare, and policymaking, where it enhances trust, transparency, and accessibility in hypothesis validation. By making model behavior more interpretable, explainability techniques support a clearer understanding and communication of computational reasoning. They also promote scientific innovation by revealing hidden patterns within hypotheses that can lead to deeper insights. However, challenges persist in maintaining a balance between interpretability and model performance, as overly simplified explanations may obscure the complexity of advanced hypotheses. Moreover, existing explainability tools often struggle with abstract or unconventional hypotheses generated by models that lack well-defined decision boundaries, limiting their applicability in frontier scientific contexts. Despite these limitations, explainability and interpretability remain foundational for promoting scientific rigor, especially as hypothesis generation becomes more data-driven and automated. Ensuring transparent and interpretable validation will be essential for bridging the gap between complex computational reasoning and actionable, real-world decision-making.

\subsection{Hybrid Validation Methods}
While individual validation strategies such as empirical testing, simulation, or predictive modeling are valuable independently, many scientific domains benefit most from hybrid validation methods that combine multiple approaches to enhance reliability, generalizability, and interpretability \cite{qi2024large, aubin2024llms}. Hybrid validation frameworks leverage the complementary strengths of diverse validation techniques to mitigate individual limitations and address the complex nature of scientific hypotheses.
For example, in biomedical research, hypotheses often undergo predictive validation using trained classifiers or generative models, followed by simulation-based testing in virtual biological environments \cite{laurent2024lab}. Promising hypotheses are then escalated to experimental validation under laboratory conditions \cite{schmidgall2024agentclinic}. This progressive validation pipeline balances scalability with scientific rigor and helps conserve resources by filtering out implausible hypotheses early in the process.
In materials science, hybrid frameworks often integrate simulation-driven methods with causal inference models to understand the mechanistic underpinnings of observed behaviors \cite{ghafarollahi2024sciagents}. High throughput simulations generate candidate hypotheses, while statistical methods validate causal relations among physical properties. These insights are subsequently verified through real-world experiments, enabling both theoretical insight and empirical confidence.
Furthermore, human AI collaborative validation is frequently combined with benchmarking and explainability techniques. For instance, hypotheses generated by language models can be benchmarked against curated datasets or domain-specific baselines, and their justifications can be interpreted using explainable artificial intelligence methods \cite{shavit2023practices}. Domain experts review these explanations to ensure alignment with established field standards. This hybrid setup ensures that hypotheses are computationally viable, epistemically sound, and ethically aligned \cite{jaradeh2019open}.

Hybrid validation methods also support interdisciplinary hypothesis evaluation, where no single domain has sufficient ground truth or validation infrastructure. In such settings, modular validation architectures activate specialized components for different subdomains, such as simulations for physics, knowledge graphs for biology, and crowd-based assessment for sociotechnical hypotheses while maintaining a coherent end-to-end evaluation pipeline \cite{zhou2024hypothesisgenerationlargelanguage, sybrandt2017moliere}.
Finally, hybrid validation facilitates risk-aware hypothesis evaluation. High-risk, high-impact ideas can be evaluated using risk-weighted scoring functions that combine quantitative metrics such as predictive accuracy and novelty with qualitative assessments such as feasibility narratives obtained from expert feedback \cite{fok2024search, wang2023scimon}. These approaches are especially critical in frontier scientific research, where uncertainty is high and exploratory validation is necessary.

In summary, hybrid validation methods offer a pragmatic and flexible approach to hypothesis evaluation by synthesizing diverse validation signals into a coherent judgment framework. As scientific discovery becomes increasingly interdisciplinary and automated, hybrid validation pipelines will be essential for balancing scalability, trust, and scientific robustness.

\textbf{Conclusion.}
The categorization of hypothesis validation methodologies highlights their diverse applicability across scientific and technological domains. Each approach, from the \textit{precision and empirical rigor of experimental validation} to the \textit{scalability and efficiency of predictive and simulation-driven models}, plays a distinct role in ensuring the robustness of validated hypotheses. \textit{Benchmarking and standardized testing} provide structured comparability, fostering reproducibility and competitive assessment, while \textit{causal inference validation} uncovers mechanistic relationships beyond correlations, strengthening the explanatory power of scientific claims.  
The integration of \textit{human-AI collaboration and crowdsourced validation} enhances interpretability and accessibility, ensuring that hypotheses are not only statistically sound but also contextually meaningful. \textit{Cross-disciplinary and multi-agent validation} extend these capabilities by enabling adaptability and scalability, allowing knowledge transfer across diverse domains and leveraging distributed intelligence for large-scale hypothesis testing. Furthermore, \textit{explainability-focused validation approaches} enhance transparency and trust, bridging the gap between complex computational models and human decision-making.  

By strategically combining these methodologies, researchers can construct \textit{robust, scalable, and adaptive validation frameworks} tailored to modern, data-driven research challenges. This convergence of methodologies not only accelerates scientific discovery but also fosters \textit{innovation, interdisciplinary breakthroughs, and higher standards of reliability and generalizability in hypothesis validation}, ensuring that research remains both impactful and reproducible in an ever-evolving scientific landscape.
\section{Future Directions: Overcoming Limitations in Hypothesis Creation and Validation}
\label{sec:future_directions}

\begin{figure*}
    \centering
    \includegraphics[width=\linewidth]{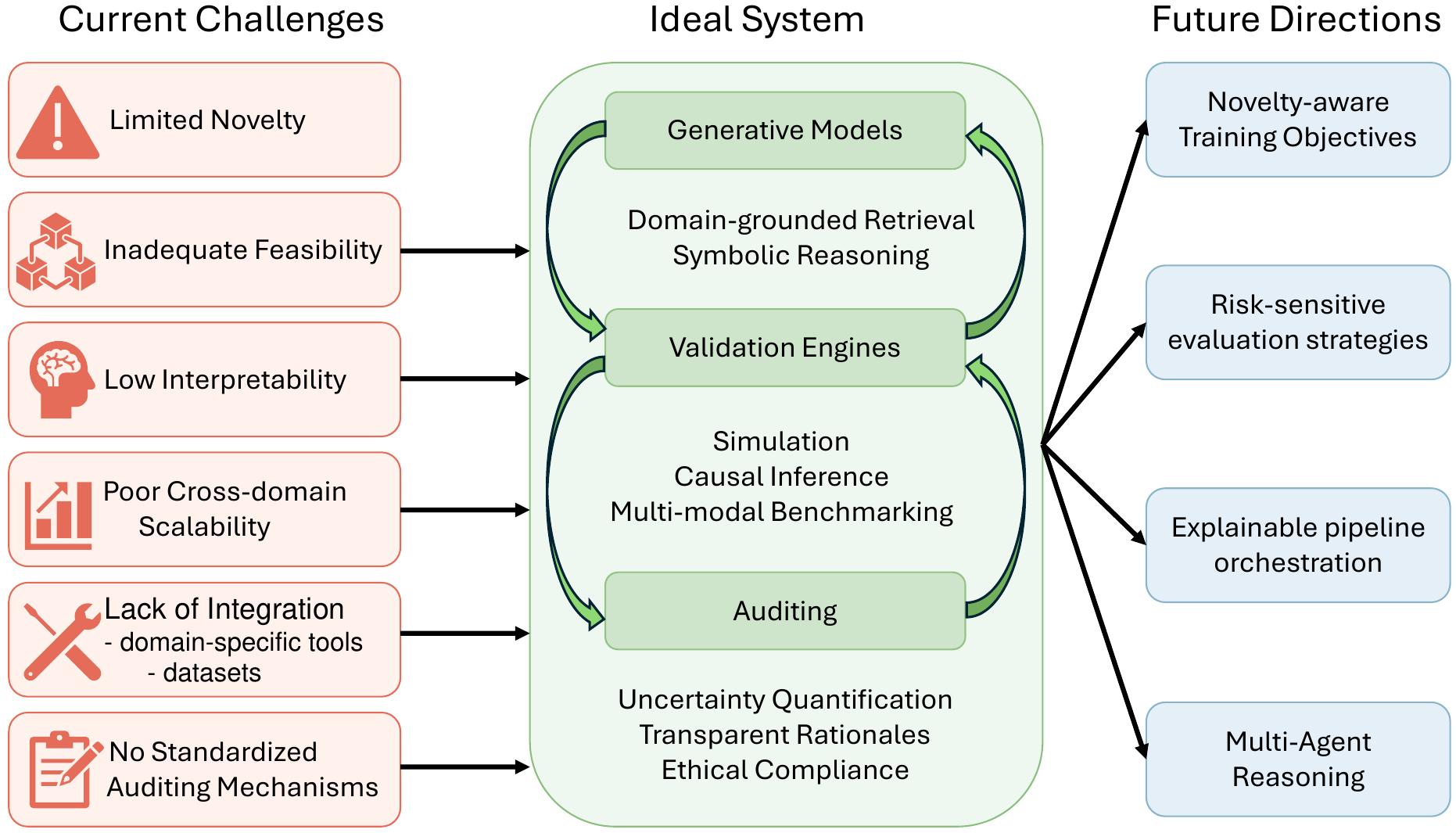}
    \caption{Roadmap for future directions in LLM-based scientific hypothesis generation and validation. The figure connects current challenges, such as limited novelty, feasibility issues, and lack of interpretability, to future directions like novelty-aware training objectives, risk-sensitive evaluation strategies, explainable pipeline orchestration, and multi-agent reasoning. The figure outlines components of an ideal system integrating generative models, validation engines, and auditing.}
    \label{fig:figure_5}
\end{figure*}

The transformative potential of LLMs in scientific hypothesis generation and validation is undeniable. However, their current capabilities are constrained by challenges such as limited novelty generation, feasibility assessment, pervasive data biases, inadequate interpretability, scalability, and adaptability. Addressing these limitations is essential to fully harness their potential across diverse scientific domains.
While LLM-driven advancements in hypothesis generation and validation have shown significant promise, the complexity and interdisciplinary demands of modern research require solutions that transcend existing limitations. Realizing the potential of LLMs to address high-risk, cross-disciplinary challenges demands a focused effort to overcome technical and methodological barriers, ensuring their integration into diverse scientific workflows. Figure ~\ref{fig:figure_5} provides a future roadmap connecting current limitations in LLM-driven hypothesis generation and validation with targeted capabilities needed in next-generation systems to achieve an ideal system.

This section highlights key challenges faced by LLM-based systems and presents actionable strategies to address them. By exploring emerging methodologies and identifying opportunities for innovation, we aim to provide a strategic roadmap for advancing hypothesis-driven discovery. These efforts will help make LLMs more robust, adaptable, and impactful tools for fostering scientific breakthroughs and addressing pressing global challenges.

\subsection{Enhancing Novelty in Hypothesis Creation}
Large language models (LLMs) are often constrained by their reliance on established knowledge bases and historical data, which can limit their ability to produce transformative or genuinely novel hypotheses. While this grounding ensures that hypotheses align with current scientific understanding, it also risks reinforcing conventional paradigms. Such reliance on past knowledge restricts the exploration of counterintuitive or uncharted areas, particularly in fields like theoretical sciences or frontier technologies that demand innovative thinking. Such reliance inadvertently reinforces existing biases and hinders transformative innovation.
These limitations have several implications. They restrict the potential for paradigm-shifting discoveries, often favoring incremental advancements over groundbreaking ideas. Furthermore, fields that require unconventional or speculative approaches remain underserved, and underrepresented or unconventional research areas risk being overlooked, perpetuating existing biases in scientific exploration.

\textbf{Strategies for Overcoming Limitations:}
\begin{enumerate}
    \item \textbf{Incorporating Generative Exploration Models:}  
    Utilize creative frameworks, such as generative adversarial networks (GANs) or reinforcement learning, to introduce variability and randomness into hypothesis creation.  
    This approach encourages models to challenge established norms and explore unconventional ideas.

    \item \textbf{Reducing Historical Bias:}  
    Apply techniques like data augmentation and debiasing~\cite{bolukbasi2016man, liang2020towards, dhamala2021bold} to broaden the scope of input datasets, thereby reducing the overfitting of models to historical patterns.  
    By diversifying inputs, models are more likely to generate hypotheses that go beyond traditional paradigms.

    \item \textbf{Enhancing Human-AI Collaboration:}  
    Develop hybrid systems where LLMs propose unconventional hypotheses, which are refined and validated through iterative feedback from domain experts.  
    This collaborative approach combines AI’s ability to explore vast hypothesis spaces with human expertise in assessing practicality and relevance.

    \item \textbf{Promoting Cross-Domain Knowledge Transfer:}  
    Integrate insights from unrelated fields, such as applying AI methodologies to materials science or borrowing biological frameworks for AI development.  
    The convergence of knowledge from multiple disciplines often results in unexpected connections and innovative hypotheses.

    \item \textbf{Establishing Novelty Metrics:}  
    Introduce specific metrics to evaluate the novelty of generated hypotheses, such as divergence from existing literature or exploration of uncharted parameter spaces.  
    Explicit metrics ensure that models prioritize innovation and generate hypotheses that challenge existing knowledge.
\end{enumerate}

\subsection{Improving Feasibility Assessments}
Hypothesis validation methods often rely heavily on computational proxies or simulated environments to evaluate feasibility. While these approaches provide scalability and cost efficiency, they frequently fall short of aligning with real-world experimental outcomes. This disconnect is particularly problematic in domains like biomedicine and materials science, where empirical validation is not only preferred but often essential for establishing the reliability of generated hypotheses. The absence of robust empirical validation mechanisms undermines the credibility of computationally derived hypotheses, creating a gap between theoretical predictions and practical applicability.

These challenges carry significant implications. Discrepancies between computational predictions and experimental results erode trust in the reliability of hypotheses, limiting their adoption in critical fields. Domains such as drug development and structural engineering, which demand rigorous empirical evidence, face particular constraints in leveraging computational validation approaches. Furthermore, the lack of alignment with experimental outcomes slows progress in high-stakes applications where feasibility is a prerequisite for real-world deployment. Addressing these limitations is essential for ensuring that hypothesis validation systems fulfill their potential to drive impactful and trustworthy scientific advancements.

\textbf{Strategies for Overcoming Limitations:}
\begin{enumerate}
    \item \textbf{Integrating Experimental Validation Pipelines:}  
    Develop frameworks that link computational predictions with empirical validation methods, such as automated laboratory systems or experimental datasets.  
    This integration enhances reliability by directly connecting computational outputs to real-world results.

    \item \textbf{Establishing Iterative Feedback Mechanisms:}  
    Create feedback loops where experimental outcomes are used to iteratively refine computational models.  
    These mechanisms ensure models continuously adapt to real-world constraints, improving their practical relevance.

    \item \textbf{Adopting Sim-to-Real Transfer Techniques:}  
    Employ methodologies like domain randomization to bridge the gap between simulated environments and experimental realities.  
    These techniques enhance the robustness and applicability of computational predictions in real-world settings.

    \item \textbf{Developing Feasibility-Centric Metrics:}  
    Introduce metrics specifically designed to quantify feasibility, such as experimental reproducibility scores or validation success rates.  
    Such metrics provide a systematic way to assess the practical applicability of hypotheses.

    \item \textbf{Fostering Interdisciplinary Collaborations:}  
    Promote partnerships between computational and experimental researchers to ground hypotheses in practical constraints.  
    These collaborations leverage diverse expertise to enhance the feasibility and applicability of generated hypotheses.
\end{enumerate}

\subsection{Addressing Domain-Specific and Interdisciplinary Challenges}

Hypothesis creation and validation approaches frequently encounter significant challenges in addressing the specialized needs of distinct scientific domains. Each domain often presents unique data structures, terminologies, and methodologies that must be accounted for to ensure accurate and meaningful outcomes. These challenges are further compounded in interdisciplinary research, where integrating diverse knowledge bases and reconciling inconsistent standards across fields is critical. For example, the synthesis of data from biology and artificial intelligence demands a balance between biological intricacies and computational frameworks.

The implications of these challenges are profound. In highly specialized fields like synthetic biology or astrophysics, hypothesis generation approaches may struggle to accommodate domain-specific complexities, thereby reducing their effectiveness. Similarly, interdisciplinary collaboration is often hindered by mismatched methodologies and datasets, slowing progress in areas that rely on cross-domain insights. Emerging fields like bioinformatics or AI-driven materials science exemplify the importance of seamless integration, as the transferability of insights across disciplines is crucial for innovation. Addressing these limitations is essential for advancing hypothesis creation and validation systems that are both domain-sensitive and interdisciplinary in scope.

\textbf{Strategies for Overcoming Limitations:}
\begin{enumerate}
    \item \textbf{Domain-Specific Fine-Tuning:}  
    Tailor hypothesis generation models to specific domains using curated datasets and terminologies.  
    This approach enhances model relevance and accuracy by aligning with the standards of the targeted field.

    \item \textbf{Cross-Disciplinary Ontologies:}  
    Create standardized frameworks and shared ontologies to integrate knowledge across fields.  
    This solution bridges terminological and methodological gaps, enabling seamless interdisciplinary hypothesis generation.

    \item \textbf{Knowledge Graphs for Interdisciplinary Research:}  
    Leverage knowledge graphs to connect domain-specific concepts and uncover potential synergies between fields.  
    These graphs provide a structured framework for exploring cross-disciplinary relationships.

    \item \textbf{Collaborative Platforms:}  
    Develop platforms that enable researchers from diverse domains to co-create, refine, and validate hypotheses collaboratively.  
    These platforms facilitate communication and integration of expertise, reducing barriers to interdisciplinary innovation.

    \item \textbf{Hybrid Models Combining Domain Expertise and General Knowledge:}  
    Integrate domain-specific models with general-purpose LLMs to capitalize on both specialized and broad knowledge.  
    This hybrid approach ensures hypotheses are rigorous and grounded while benefiting from interdisciplinary insights.
\end{enumerate}

\subsection{Mitigating Data Limitations and Biases}

Many hypothesis creation and validation approaches depend heavily on datasets that are often biased, incomplete, or fail to represent the complexities of emerging scientific fields. These shortcomings significantly impact the diversity and generalizability of the generated hypotheses, limiting their applicability across underrepresented areas or demographics. Additionally, biases inherent in training data risk perpetuating systemic inequities, which can undermine trust in AI-driven discoveries and hinder the adoption of these technologies in critical applications.

The implications of these limitations are far-reaching. Hypotheses that are not inclusive of diverse datasets may fail to address the unique challenges of specific populations or regions, limiting their real-world utility. Furthermore, systemic biases introduced during hypothesis creation can erode confidence in the fairness and reliability of scientific discoveries. In emerging fields, where high-quality data is often fragmented or scarce, these issues constrain innovation and slow the progress of groundbreaking research. Addressing these challenges is essential to ensure that hypothesis creation and validation systems promote equity, inclusivity, and the advancement of science in all domains.

\textbf{Strategies for Overcoming Limitations:}
\begin{enumerate}
    \item \textbf{Data Augmentation Techniques:}  
    Apply methods such as synthetic data generation, cross-domain transfer, or oversampling to enrich datasets in underrepresented areas.  
    Augmented datasets improve diversity and enable comprehensive hypothesis exploration.

    \item \textbf{Bias Detection and Mitigation Frameworks:}  
    Implement tools that identify and correct biases, such as fairness metrics or adversarial debiasing.  
    These frameworks ensure equitable and unbiased hypothesis generation.

    \item \textbf{Collaborative Dataset Curation:}  
    Promote interdisciplinary and community-driven efforts to curate diverse, high-quality datasets.  
    Collaborative curation reduces blind spots and enhances dataset coverage.

    \item \textbf{Active Learning Approaches:}  
    Use active learning to iteratively refine datasets by prioritizing samples that maximize model performance.  
    This strategy optimizes data collection and improves dataset representativeness.

    \item \textbf{Openness and Transparency in Dataset Usage:}  
    Advocate for open-access repositories and standardized metadata to detail dataset limitations and biases.  
    Transparency fosters accountability and helps users interpret results in context.
\end{enumerate}

\subsection{Enhancing Interpretability and Transparency}

AI-based hypothesis creation and validation models frequently lack interpretability and transparency, posing significant challenges to their adoption and utility. The intricate nature of these models often obscures the logic and reasoning behind the generated hypotheses, making it difficult for researchers to assess their validity and trustworthiness. This opacity not only limits the understanding of the underlying processes but also creates barriers to effective validation and refinement by domain experts.

The implications of these challenges are profound. A lack of transparency reduces trust in AI-driven hypotheses, particularly in high-stakes areas like healthcare and public policy, where decisions can have far-reaching consequences. Furthermore, domain experts may struggle to engage meaningfully with these systems, as opaque methodologies limit their ability to critically evaluate and refine hypotheses. This lack of clarity also hinders interdisciplinary collaboration, as researchers from diverse fields may find it difficult to reconcile opaque AI-generated results with established domain-specific practices. Enhancing interpretability and transparency in AI models is therefore essential for fostering trust, enabling effective validation, and promoting collaborative advancements in hypothesis-driven research.

\textbf{Strategies for Overcoming Limitations:}
\begin{enumerate}
    \item \textbf{Adoption of Explainable AI (XAI) Techniques:}  
    Incorporate tools like saliency maps, SHAP (SHapley Additive exPlanations), and LIME (Local Interpretable Model-Agnostic Explanations) to clarify AI-generated hypotheses.  
    These techniques improve user understanding and trust.

    \item \textbf{Human-Readable Outputs:}  
    Design models to output hypotheses in interpretable formats, such as graphs, natural language summaries, or decision trees.  
    Clear outputs facilitate hypothesis validation and enhance communication among researchers.

    \item \textbf{Model Auditing and Validation:}  
    Establish auditing frameworks to evaluate the transparency of model predictions and hypothesis generation processes.  
    Auditing promotes accountability and identifies inaccuracies in reasoning.

    \item \textbf{Interactive Visualization Tools:}  
    Develop platforms that allow users to explore model-generated hypotheses and underlying data visually.  
    Visualization tools help identify patterns, anomalies, and areas needing further investigation.

    \item \textbf{Open-Sourcing Models and Workflows:}  
    Support open-source development of AI models and validation pipelines to enable independent scrutiny and reproducibility.  
    Open access increases transparency and fosters interdisciplinary collaboration.
\end{enumerate}

\subsection{Encouraging High-Risk, High-Reward Hypotheses}

Current frameworks for hypothesis creation and validation frequently emphasize safety and reliability, which, while valuable, often discourage the pursuit of high-risk, high-reward hypotheses. By prioritizing incremental progress and aligning closely with established paradigms, these frameworks inadvertently constrain the exploration of transformative ideas that challenge conventional norms. This cautious approach limits the potential to venture into uncharted territories where groundbreaking discoveries might emerge.

The implications of this risk-averse mindset are significant. It restricts the scientific community's ability to achieve paradigm-shifting advancements, particularly in fields where unconventional thinking is essential. Researchers may be deterred from exploring novel or controversial ideas due to the perceived risks and lack of institutional support, further reinforcing existing knowledge silos. This focus on low-risk, incremental progress slows innovation, particularly in emerging fields where disruptive breakthroughs are most needed. Developing frameworks that balance reliability with the freedom to explore bold hypotheses is critical for driving transformative scientific and technological advancements.

\textbf{Strategies for Overcoming Limitations:}
\begin{enumerate}
    \item \textbf{Incorporating Risk-Tolerant Metrics:}  
    Introduce evaluation metrics that reward hypotheses with high potential impact despite their inherent uncertainty.  
    This approach fosters exploration of transformative ideas while maintaining a balance with feasibility.

    \item \textbf{Exploratory Funding Models:}  
    Advocate for funding mechanisms, such as "moonshot" grants, to support high-risk, high-reward research.  
    Dedicated financial support reduces barriers and incentivizes bold innovations.

    \item \textbf{Scenario-Based Validation Frameworks:}  
    Create frameworks that assess hypotheses across diverse scenarios to evaluate long-term risks and rewards.  
    Scenario-based evaluations provide a nuanced understanding of high-risk ideas.

    \item \textbf{Cross-Disciplinary Teams for Validation:}  
    Involve interdisciplinary teams to evaluate unconventional hypotheses from multiple perspectives.  
    Diverse expertise ensures a balanced assessment of high-risk proposals.

    \item \textbf{Recognition and Incentive Structures:}  
    Establish systems that reward researchers for pursuing innovative ideas, regardless of immediate success.  
    Positive incentives promote a culture of creativity and exploration.
\end{enumerate}

\subsection{Scalability in Data Integration}

Hypothesis generation and validation approaches often encounter significant scalability challenges when tasked with integrating large-scale, heterogeneous datasets. As the volume and diversity of available data continue to grow, computational demands increase exponentially, complicating the effective processing and analysis of these datasets. The complexity of managing such multidimensional, real-world data further exacerbates the difficulty of deriving meaningful insights.

These scalability limitations have notable implications for scientific discovery. They hinder the ability to generate insights that leverage information from multiple domains or large-scale datasets, reducing the potential for cross-disciplinary innovation. Additionally, the increased computational costs and decreased efficiency associated with handling vast datasets strain hypothesis workflows, making them less practical for widespread application. Without addressing these challenges, current approaches risk becoming obsolete in the face of the ever-expanding data landscape, underscoring the need for more scalable and efficient solutions.

\textbf{Strategies for Overcoming Limitations:}
\begin{enumerate}
    \item \textbf{Scalable Data Processing Architectures:}  
    Utilize distributed computing frameworks, such as Apache Spark or TensorFlow Extended, to handle large-scale datasets.  
    Parallel processing reduces bottlenecks in data-intensive tasks.

    \item \textbf{Federated Learning for Data Integration:}  
    Implement federated learning to integrate distributed data sources without centralizing them.  
    This preserves privacy while enabling scalable analysis across systems.

    \item \textbf{Advanced Data Preprocessing Pipelines:}  
    Develop automated workflows for normalizing, cleaning, and aligning heterogeneous datasets.  
    Preprocessing ensures data quality and consistency for improved hypothesis generation.

    \item \textbf{Knowledge Graphs for Data Fusion:}  
    Use knowledge graphs to unify datasets by representing relationships across sources.  
    Semantic linking facilitates the discovery of hidden insights and relationships.

    \item \textbf{Optimized Query and Storage Systems:}  
    Deploy systems like graph databases or NoSQL solutions to manage multidimensional data efficiently.  
    These tools enhance storage and retrieval capabilities, improving real-time integration.
\end{enumerate}

\subsection{Feedback-Driven Refinement}

Hypothesis creation and validation systems frequently lack robust feedback mechanisms that enable the integration of iterative expert input. This absence of interaction diminishes the domain-specific relevance of generated hypotheses, as these systems are unable to benefit from nuanced corrections or refinements provided by subject matter experts. Furthermore, without continuous feedback loops, these models struggle to adapt effectively to evolving research priorities or new datasets, limiting their overall utility.

The implications of this limitation are significant. The exclusion of expert insights reduces the quality and contextual accuracy of hypotheses, often leading to outputs that are less applicable or meaningful in real-world scenarios. Additionally, the inability to identify and correct errors in AI-generated hypotheses undermines trust in these systems and restricts their adoption in critical fields. To remain relevant and impactful, hypothesis creation and validation frameworks must prioritize the development of adaptable, expert-informed feedback mechanisms that align with dynamic research needs.

\textbf{Strategies for Overcoming Limitations:}
\begin{enumerate}
    \item \textbf{Human-in-the-Loop Systems:}  
    Design systems that allow experts to refine hypotheses iteratively through feedback loops.  
    This ensures hypotheses align with domain-specific standards and relevance.

    \item \textbf{Interactive Model Interfaces:}  
    Create user-friendly interfaces for visualizing, modifying, and providing feedback on hypotheses.  
    Interactive tools enhance transparency and facilitate dynamic refinement.

    \item \textbf{Reinforcement Learning with Human Feedback:}  
    Use reinforcement learning algorithms to optimize models based on expert feedback.  
    Continuous improvement adapts models to evolving research needs.

    \item \textbf{Iterative Validation Pipelines:}  
    Develop pipelines where hypotheses undergo cycles of AI validation and expert evaluation.  
    Iterative approaches improve robustness and minimize errors.

    \item \textbf{Crowdsourced Feedback Mechanisms:}  
    Incorporate diverse insights through crowdsourcing platforms for collaborative hypothesis refinement. 
    Crowdsourcing broadens perspectives and enhances inclusivity in research processes.
\end{enumerate}

Addressing limitations in hypothesis creation and validation is crucial for realizing the full potential of LLMs in scientific discovery. By fostering novelty, improving feasibility, and enhancing interpretability, these tools can become robust enablers of transformative research.
Tackling challenges like data biases, scalability, and interdisciplinary integration will empower researchers to drive breakthroughs across diverse fields. Strategies such as interdisciplinary frameworks, human-in-the-loop systems, and novel evaluation metrics are essential for encouraging bold, high-impact hypotheses.
Implementing these advancements will bridge existing gaps and position LLMs as indispensable allies in accelerating innovation and fostering a new era of scientific discovery.
\section{Conclusion}
\label{sec:conclusion}

This survey has explored the transformative role of Large Language Models (LLMs) in scientific hypothesis creation and validation, examining their methodologies, datasets, limitations, and future directions. LLMs exhibit exceptional capabilities in processing vast datasets, identifying patterns, and generating insights, leveraging approaches such as knowledge graphs, retrieval-augmented generation, predictive modeling, and simulation-based validation. These methods offer diverse strengths, yet they remain constrained by challenges such as limited novelty, data biases, domain-specific constraints, and the reliance on computational proxies rather than empirical validation.
To fully harness the potential of LLMs in scientific discovery, future research must focus on advancing novelty, feasibility, and interdisciplinary integration. Enhancing novelty through advanced generative exploration frameworks, such as generative adversarial models and contrastive learning, can help overcome the tendency of LLMs to rely on existing knowledge. Strengthening feasibility via hybrid computational-empirical pipelines will ensure that hypotheses generated by LLMs align with experimental validation. Additionally, addressing interdisciplinary challenges through cross-domain ontologies, federated learning, and collaborative AI-human interfaces will enhance the adaptability of LLMs in complex research landscapes.

Ensuring data integrity and mitigating biases is another critical step. Techniques including strategic knowledge graph augmentation, adversarial debiasing methods, and fairness-aware machine learning models can help mitigate biases and enhance the scalability, inclusivity, and trustworthiness of LLM-driven hypothesis generation. Moreover, interpretability-focused systems, incorporating explainable AI techniques and uncertainty quantification, will be essential for bridging the gap between computational insights and human intuition.
The future of LLM-driven scientific discovery lies in iterative human-AI collaboration, where researchers and AI systems co-develop hypotheses in a risk-tolerant and adaptive framework. By integrating explainability, risk assessment, and interdisciplinary synthesis, LLMs can serve as powerful catalysts for high-risk, high-reward research, accelerating breakthroughs in biomedicine, materials science, climate modeling, and beyond.
In conclusion, as advancements in LLM technology continue to evolve, their role in hypothesis-driven research will expand, redefining scientific exploration. By addressing existing limitations and fostering a collaborative, interpretability-driven ecosystem, LLMs hold the potential to usher in a new era of transformative knowledge creation, interdisciplinary discovery, and scalable scientific innovation.

\bibliographystyle{unsrtnat}
\bibliography{references}

\end{document}